\documentclass[journal]{IEEEtran} 
\usepackage{amsmath,amsfonts}
\usepackage{amssymb}
\usepackage{algorithmic}
\usepackage{algorithm}
\usepackage{array}
\usepackage[caption=false,font=normalsize,labelfont=sf,textfont=sf]{subfig}
\usepackage{textcomp}
\usepackage{stfloats}
\usepackage{url}
\usepackage{verbatim}
\usepackage{graphicx}
\usepackage{cite}
\usepackage{wrapfig,lipsum,booktabs}
\usepackage{multirow}
\usepackage[dvipsnames]{xcolor}
\usepackage{hyperref}
\usepackage{todonotes}
\newif\ifrevisions
\revisionstrue   % <-- comment out or remove to hide revision-only bits

\newcommand{\trorev}[1]{%
  \ifrevisions
    #1%
  \fi
}

\newcommand{\troanony}[1]{%
  \ifrevisions
  \else
    #1%
  \fi
}

\begin{document}

\title{TensorTouch: Calibration of Tactile Sensors for High Resolution Stress Tensor and Deformation for Dexterous Manipulation}

% \title{TensorTouch: High Resolution Stress Tensor and Deformation Estimation of 3D Optical Tactile Sensors for Dexterous Manipulation}

% calibrating 3D optical tactile sensor for high resolution stress tensor and deformation towards advanced dexterity
% stress tensor synonym ? 
% TensorTouch: Calibrated Stress Tensor and Contact Force Estimation from 3D Optical Tactile Sensors for Advanced Dexterous Manipulation

% TensorTouch: High Resolution Calibrated Stress Tensor Estimation of 3D Optical Tactile Sensors for Advanced Dexterous Manipulation

% densetact in title? 

% calibration in title / benefit of calibrating 
% 1. specific knowledge of force, then we can measure that for safety check 
% 2. translating btw different embodiment - glove / inductor sensor , etc. skill transfer on imitation learning / cross embodiment  regardless of type of tactile sensor 

% Calibrating Optical Tactile Sensors for Stress Field Reconstruction and Dexterous Manipulation

% Physics-Based Stress Tensor Estimation for Optical Tactile Sensors

% High-Resolution Stress Tensor Estimation for Vision-Based Tactile Sensors

% Generalizable Stress Tensor Estimation for Optical Tactile Sensors with Large Deformations

\author{\trorev{Won Kyung Do$^{1}$, Matthew Strong$^{2}$, Aiden Swann$^{1}$, Boshu Lei$^{3}$, Monroe Kennedy III$^{1,2}$
\thanks{This research was supported by NSF Graduate Research Fellowship No.
DGE-2146755 and NSF Grant No. 2142773, 2220867.}% <-this % stops a space
\thanks{Project website: \href{https://tensor-touch.github.io/}{https://tensor-touch.github.io/}.}
\thanks{$[\cdot]^{1}$ are with the Department of Mechanical Engineering,
$[\cdot]^{2}$ is with Department of Computer Science,
        Stanford University, Stanford CA, USA and $[\cdot]^{3}$ is with School of Engineering and Applied Science,
        University of Pennsylvania, Philadelphia PA, USA.
        Emails: \{wkdo, mastro1, swann, monroek\}@stanford.edu, leiboshu@seas.upenn.edu}}
        \troanony{Anonymous 
\thanks{All the funding information and author information are hidden.}
}}

% \markboth{IEEE TRANSACTIONS ON ROBOTICS,~Vol.~XX, No.~X, XXXXX~20XX}%


\maketitle

\begin{abstract}
% Advanced dexterous manipulation involving multiple simultaneous contacts across different surfaces, like pinching coins from ground or manipulating intertwined objects, remains challenging for robotic systems. Such tasks exceed the capabilities of vision and proprioception alone, requiring high-resolution tactile sensing with calibrated physical metrics. Raw optical tactile sensor images, while information-rich, lack interpretability and cross-sensor transferability, limiting their utility in real-world applications. TensorTouch addresses this challenge by integrating finite element analysis with deep learning to extract comprehensive contact information from optical tactile sensors, including stress tensors, deformation fields, and force distributions at pixel-level resolution. Our framework achieves sub-millimeter position accuracy and precise force estimation while supporting large sensor deformations, crucial for manipulating soft or delicate objects. Experimental validation demonstrates the effectiveness of our method in a challenging multi-object manipulation task of selectively grasping one of two strings based on detected motion-achieving 90\% success with objects of different material properties. This enables new capabilities in contact-rich manipulation scenarios previously inaccessible to robotic systems.

Advanced dexterous manipulation involving multiple simultaneous contacts across different surfaces, like pinching coins from ground or manipulating intertwined objects, remains challenging for robotic systems. Such tasks exceed the capabilities of vision and proprioception alone, requiring high-resolution tactile sensing with calibrated physical metrics. Raw optical tactile sensor images, while information-rich, lack interpretability and cross-sensor transferability, limiting their real-world utility. TensorTouch addresses this challenge by integrating finite element analysis with deep learning to extract comprehensive contact information from optical tactile sensors, including stress tensors, deformation fields, and force distributions at pixel-level resolution. The TensorTouch framework achieves sub-millimeter position accuracy and precise force estimation while supporting large sensor deformations crucial for manipulating soft objects. Experimental validation demonstrates 90\% success in selectively grasping one of two strings based on detected motion, enabling new contact-rich manipulation capabilities previously inaccessible to robotic systems.
\end{abstract}


\section{Introduction}

Dexterous manipulation in contact-rich environments remains as significant challenges in robotics research, with applications ranging from industrial assembly to household assistance. While significant progress has been made in simple manipulation tasks such as pick-and-place operations, folding laundry, or surface swiping through position control and vision-based feedback \cite{black2024pi0, intelligence2025pi05} or even with force control \cite{suh2025dexterous}, more complex manipulation scenarios remain challenging. These advanced tasks often require understanding and responding to multi-contact interactions with multiple objects or surfaces simultaneously.

\begin{figure}[t]
\centering
\includegraphics[width=0.95\linewidth]{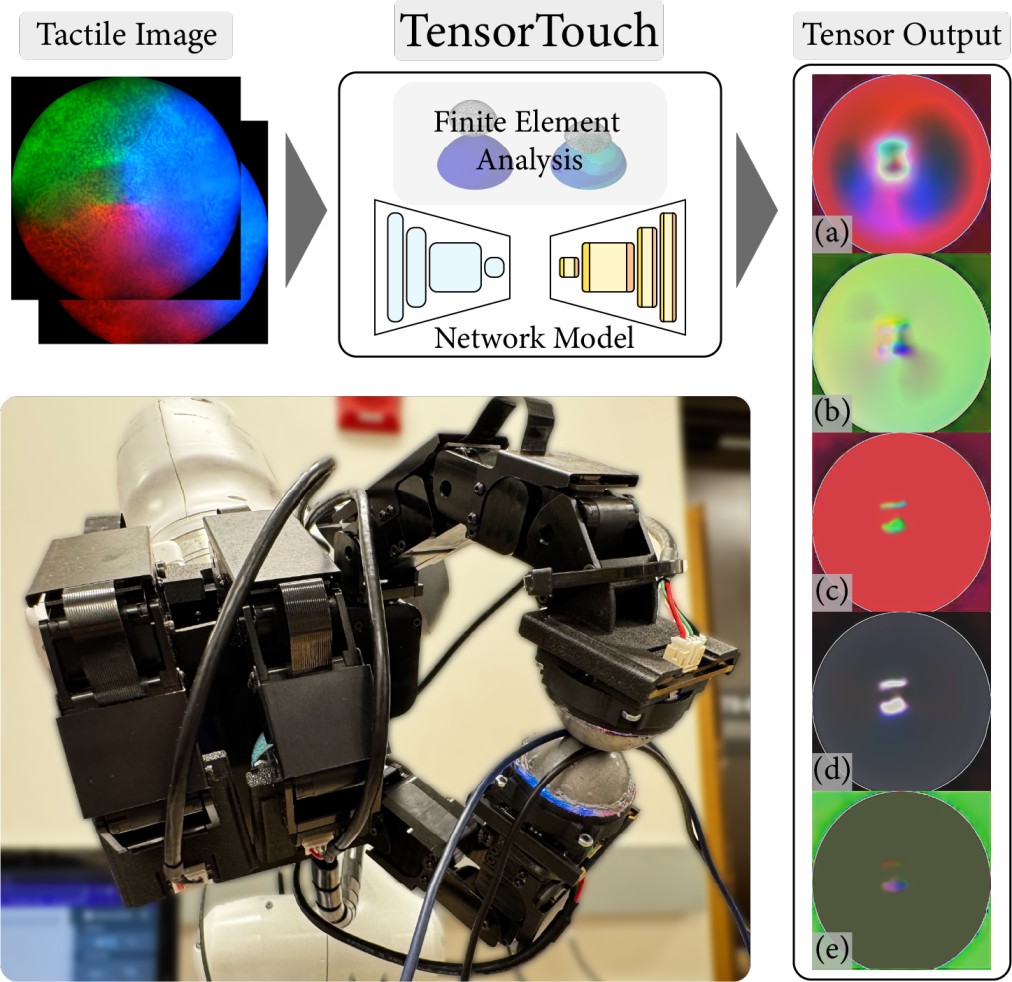}
\caption{TensorTouch framework applied to multi-finger robotic manipulation with optical tactile sensors mounted on fingertips grasping two strings. The system processes tactile images (left) through FE analysis to generate comprehensive tensor outputs: (a) displacement, (b) normal stress, (c) shear stress, (d) contact normal force, and (e) contact shear force.}
\vspace{-1.7em}
\label{fig:main}
\end{figure}

Advanced dexterous manipulation, as defined in this work, involves scenarios where multiple contacts across sensing modalities must be precisely monitored to capture the full context of the manipulation. For example when humans pinch a coin from a flat surface, each finger simultaneously contacts both the coin and the surface, allowing precise perception of the coin's position and establishing force closure across its edges. Such interactions, involving multiple contacts and force distributions, are pervasive in everyday tasks yet remain difficult for robotic systems to execute reliably.

Addressing these challenges requires integrating enhanced manipulability with sophisticated sensing modalities. Specifically, multi-fingered grippers equipped with rich tactile sensing capabilities offer promising solutions for complex manipulation scenarios \cite{lin2025pp, wang2024penspin}. Vision-based tactile sensors mounted on fingertips can provide detailed information about contact geometry, force distribution, and material properties, enabling robots to understand complex force interactions across multiple objects and environmental contacts. Furthermore, advanced dexterous tasks, such as handling fragile objects and contact-rich interactions with slip and shear, often require precise estimation of shear and normal force distribution and deformation.

However, the raw output from optical tactile sensors---typically images---requires substantial processing to extract actionable information. While recent research has demonstrated the utility of simplified tactile features such as contact position or force direction for specific manipulation tasks, these approaches fail to exploit the full richness of information contained in tactile images, such as deformation and force. Moreover, relying solely on raw tactile images for policy training creates significant limitations for practical deployment. When policies are trained directly on unprocessed tactile images, they become tightly coupled to the specific sensor hardware used during training, making cross-sensor generalization nearly impossible. In contrast, extracting calibrated force information and stress distributions provides interpretable, physically meaningful metrics that can be transferred across different tactile sensing modalities. This calibration-based approach enables what could be considered `sensor-agnostic representation' in the sensing domain-the ability to substitute one tactile sensor for another without retraining policies, provided both can output standardized force metrics.

Such sensor-agnostic representations dramatically improve the scalability and practical utility of tactile-based manipulation policies, as robots can leverage multiple sensor types or even upgrade sensing hardware without requiring extensive retraining. Furthermore, calibrated force metrics facilitate more interpretable policy behavior and enable direct integration with analytical approaches to manipulation that rely on accurate force information. For these reasons, extracting comprehensive contact data, including stress tensors and deformation fields, would significantly enhance manipulation capabilities in complex scenarios.

A fundamental challenge in developing these capabilities is the lack of accurate physical simulators that can model the intricate force interactions between tactile sensors and contacted objects or surfaces. Although recent work has focused on creating GPU-accelerated simulators for marker-based tactile sensors, significant sim-to-real gaps persist due to inadequate physics modeling, particularly for large deformations and complex contact scenarios. These limitations hamper the development of policies that can transfer effectively to real-world applications.

To address these issues comprehensively, we present TensorTouch, a framework for stress tensor estimation from general optical tactile sensors designed for contact-rich manipulation tasks. Our approach integrates physics-based simulation with deep learning to extract detailed contact information from tactile images, enabling advanced manipulation capabilities even with substantial sensor deformations. The key contributions of this work include:
\begin{itemize}
    \item A comprehensive Finite Element analysis (FE analysis) framework for modeling multi-contact interactions and large deformations in various 3D-shaped optical tactile sensors, enabling high-spatial-resolution contact information extraction.
    \item A novel neural network architecture with a lightweight hierarchical vision transformer that efficiently maps tactile images to comprehensive stress tensors, contact forces, and deformation fields across the entire sensor surface.
    \item A motion capture-based data collection system that pairs tactile images with precise contact poses and force measurements from diverse object geometries, validated through experiments on robotic hands performing complex manipulation tasks.
\end{itemize}
Fig. \ref{fig:main} illustrates our complete framework applied to a challenging multi-object manipulation scenario, where optical tactile sensors mounted on robotic fingertips enable precise force and deformation estimation during two-string grasping tasks.

Our framework enables the estimation of diverse contact information, including contact area, normal and shear forces, deformed shape, and normal and shear stress distributions across the sensor surface, even under large deformations. This comprehensive tactile perception capability facilitates advanced dexterous manipulation in contact-rich environments beyond what was previously achievable.

The remainder of this paper is organized as follows: Section II reviews related work. Section III details our methodology, including the data collection system, FE analysis approach, and neural network architecture. Section IV presents experimental results, while Section V demonstrates the evaluation of TensorTouch. Finally, Section VI concludes with a discussion of limitations and future work.

\begin{figure*}
    \centering
    \includegraphics[width=0.92\textwidth]{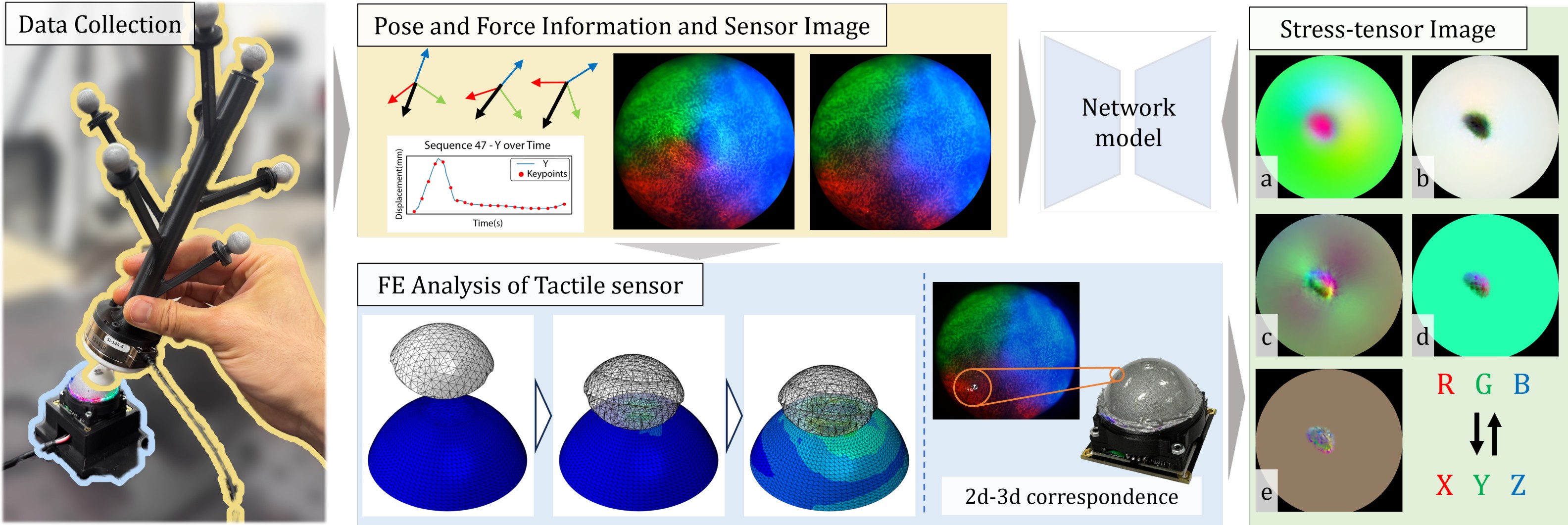}
    \caption{Overview of a pipeline. We capture real-world pose, force of indenter and sensor images 
 and simulate gel deformation and sensor motion with FE simulations. The resulting deformation and stress tensors are projected back into 2D images. Images in right part (a-e) show corresponding FE results for (a) displacement, (b) normal stress, (c) shear stress, (d) contact normal, and (e) contact shear. For each image, the R/G/B channels encode the X/Y/Z components. Finally, a deep network is trained to map sensor image pairs to stress-tensor outputs.}
 \vspace{-1.7em}
    \label{fig:main_pipeline}
\end{figure*}  
\section{Related Works}

% fin manipulation dlr \cite{kasolowsky_fine_2024, winkelbauer_learning-based_2024}
% tactile gym 2.0 \cite{lin_tactile_2022}
\subsection{Optical Tactile Sensors}

Dexterous manipulation has been proven to work better with additional modalities, especially input from touch sensors or contact information with the environment when it comes to contact-rich manipulation \cite{akinola_tacsl_2025, wang_poco_2024, qi_general_2023, kasolowsky_fine_2024, winkelbauer_learning-based_2024}. One valuable source of additional modality for dexterous manipulation is tactile information. Various types of tactile sensors are available, including those with conductive fields \cite{wettels2008biomimetic}, force sensing resistor arrays \cite{hillis1981active}, piezoelectric sensors, capacitive force sensors (either as single units or arrays), barometric sensors \cite{hou_location_2025}, and hall effect sensors \cite{Paxini2024AXGen2}.

Optical tactile sensors, which rely on vision inside deformable gel to convert high-resolution images to deformation or tactile information from the sensor surface, have been widely researched due to their ability to cover wide areas accurately with high resolution. These sensors are available in various shapes. Flat sensors, such as GelSight \cite{noauthor_gelsight_nodate}, GelSlim \cite{taylor_gelslim_2022}, DIGIT \cite{lambeta_digit_2020}, MagicTac \cite{fan_magictac_2024}, and DelTact \cite{zhang_deltact_2022}, are well-suited for parallel jaw grippers. More hemispherical shaped sensors such as DIGIT360 \cite{lambeta_digitizing_2024}, DenseTact \cite{do2022densetact}, TacTip \cite{ward-cherrier_tactip_2018}, and OmniTact \cite{padmanabha2020omnitact} offer enhanced capabilities for manipulability with various robot gripper shapes. Custom sensor designs such as GelSight Svelte \cite{zhao_gelsight_2023}, DenseTact Mini \cite{do2023dtmini}, and Insight \cite{sun_soft_2022} have been developed to further improve dexterous manipulation capabilities.

For most optical and other tactile sensors, researchers have tried to constrain the softness of the gel to achieve more accurate correspondence between force application and position measurements, which unfortunately sacrifices compliance - a critical property for certain tasks. Some sensors with alternative approaches, such as \cite{kuppuswamy_soft-bubble_2020}, successfully incorporate soft contact interaction, but the inherent size of these sensors can limit their use in dexterous manipulation. Therefore, there is a need to develop better soft deformation calibration algorithms to enable the use of softer tactile sensors while maintaining accurate measurements. 

\subsection{Requirements of Accurate Calibration for Dexterous Manipulation}

% unseen tactile representation learning, \cite{xu2025unit} directly goes to the policy training
% poco \cite{wang_poco_2024}
Recent works have proven that adding accurate tactile information to vision-based policies increases the success rate of trained policies in both reinforcement learning and imitation learning. More detailed accurate tactile representations lead to fine dexterity in both analytical ways and policy learning \cite{kasolowsky_fine_2024, winkelbauer_learning-based_2024, akinola_tacsl_2025, bronars2024texterity, xu2025unit, lin2025pp, sharma2025self}. For example, recent developments in imitation learning suggest that applying fast policy adjustments with optical flow-based force information from tactile sensors, rather than raw tactile images, on top of behavior cloning policies such as diffusion policy \cite{chi2023diffusion} provides quick reactive behavior while processing complex trajectory tracking \cite{xue2025reactive}. While raw tactile images could be an additional source of multi-modality in policy training, accurate force information from the tactile sensor reduces dimensionality and enables policies to learn tactile embeddings more easily than from raw images. Therefore, calibrating tactile sensory information is essential to achieve better and finer dexterous manipulation.

\subsection{Sensor Calibration}

Optical tactile sensors require conversion of camera images into useful information, and various approaches have been proposed for calibrating these sensors using either simulation or real-world methods. For estimating sensor deformation, GelSight-related sensors have developed an approach using Poisson equations to achieve accurate displacement estimation \cite{noauthor_gelsight_nodate}. Other approaches include binocular imaging \cite{zhang_robot_2018, cui_-hand_2022}, regression-based light modeling \cite{lin_dtact_2023}, and self-supervised neural networks \cite{sun_soft_2022} for calibrating sensors with small deformations, typically limited to 1-5mm of displacement.

However, estimating large sensor deformations requires more sophisticated approaches, such as accounting for bulging effects \cite{doDT2.0}. To address this problem more accurately, physics-based simulation analysis is recommended. Force estimation requires additional modeling beyond deformation tracking. Methods for estimating forces include tracking markers on the sensor surface \cite{ma_dense_2019} or using self-supervised neural network models \cite{sun_soft_2022, higuera_sparsh_2024} to obtain force distribution across the sensor. However, these approaches either work only with small deformations or lack the detailed modeling needed for accurate estimation during large deformations. One promising solution for estimating forces during large deformations is using physics-based simulation, specifically FE analysis (FEA), to model sensor deformation in detail.

\subsection{Sensor Calibration to FE Analysis}

For the reasons noted above, FE analysis models can provide near-ground-truth simulations for accurately calibrating a sensor's mechanical properties. FEM-based simulators enable fast analysis using physically accurate models. Since FEM modeling itself requires significant computational resources, researchers have developed more efficient alternatives including neural network models \cite{narang2021sim}, convex optimization approaches \cite{peng_3d_2024}, and simplified inverse FEM models \cite{ma_dense_2019, duong_bitac_2023, zhao_ifem20_2025} that can effectively substitute full FEA models.
For example, iFEM 2.0 \cite{zhao_ifem20_2025} utilizes a neo-Hookean model from Abaqus to successfully estimate 3D contact forces, including shear forces, for gels up to 3mm thick. However, to our knowledge, no previous work has correctly modeled large deformations of tactile sensors with varying gel softness, which limits the practical utility for handling multiple objects with various stiffness. In our work, we utilize an FE model applicable to most sensor shapes that accounts for large deformations through careful selection of the mechanical model for deformable materials, resulting in precise force and stress distribution calculations across the entire sensor.

\subsection{Learning Depth and Force Estimation from Image Input - Network Model}

Using our FE model outputs, we propose a simplified neural network-based model using a hierarchical vision transformer that estimates stress vector fields and displacements from single sensor images. Previous research has focused on estimating single 6-axis force wrenches or normal force distributions \cite{doDT2.0, ma_dense_2019, peng_3d_2024}. In contrast, our FE model provides rich force estimates including shear forces for each mesh element, enabling the complete characterization of force distribution across the entire sensor surface.
Using this detailed information, we can both construct and train deep networks  that directly estimate force distributions across the sensor surface in a pixel-wise manner. Researchers have already demonstrated the effectiveness of network models for such force estimations. For example, Sparsh \cite{higuera_sparsh_2024} introduced a self-supervised pretraining method tactile encoders, which can be effectively trained on downstream tasks, such as estimating normal and shear force distributions using optical flow in flat optical tactile sensors. Rather than relying on marker-based approaches proposed in other related works, we accurately model sensor simulation results with FEM, and use these outputs to train appropriate network architectures \cite{Hiera, singh2023effectiveness} that are significantly faster than simulation, and estimate the force distribution across the sensor surface with dense, pixel-level precision.

\section{Method} 
% \begin{itemize}
%     \item experiment setup (mocap system, hysterisis, indenter shape, ...)
%     \item Simulation (FEA simulation, keypoint extraction for simulation (simulation in real world?) )
%     \item sensor configuration (gel property, sensor design, gel shape, everything related to sensor

%     \item learning (2d-3d correspondence for 2d image preparation. pretraining including blender, model structure, )
%     \item evaluation (controlling from the displacement / shear / contact force output using 
% \end{itemize}

% \subsection{Experiment setup }
% \edit{Need to explain the setup as well has how we correspond the real world and the simulation? in here ? }
\begin{figure}[t]
\centering
\includegraphics[width=0.95\linewidth]{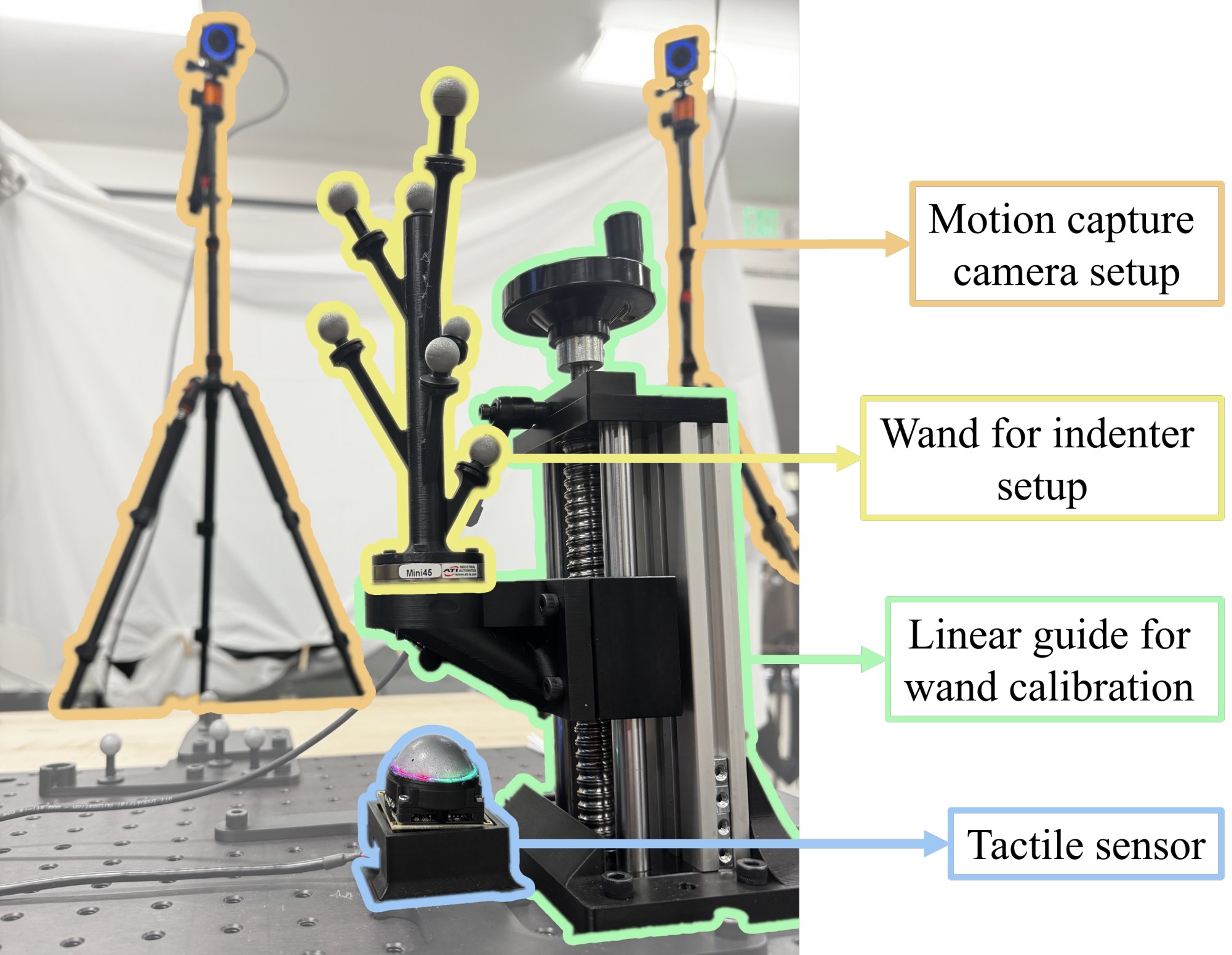}
\caption{Experimental setup for data collection with motion capture cameras (orange) and reflective markers (yellow) for indentation. The linear guide system (green) enables precise calibration of the wand position, while the tactile sensor (blue) is mounted on an optical breadboard to ensure stability.}
\vspace{-2em}
\label{fig:mocap_setup}
\end{figure}

\subsection{Motion Capture-Based Setup for Data Collection}
We developed an experimental setup for collecting comprehensive datasets from vision-based tactile sensors while supporting multiple simultaneous contacts with diverse indenter shapes. This setup is crucial for accurate sensor simulation and ensuring generalizability across different contact scenarios beyond single-point force measurements.

Our system was designed to meet critical requirements: generate random 6D poses safely within the workspace, accommodate multiple sensor form factors, simultaneously capture pose and force/torque measurements, ensure safe data collection, handle multiple simultaneous indentations, and provide cost-effective real-time capabilities with sufficient accuracy.

After evaluating robotic arms and CNC machines, we selected a motion capture-based approach for several compelling reasons. Motion capture systems provide greater freedom in generating random 6D poses compared to robotic end effectors or portable CNC machines, which are typically limited to 4 degrees of freedom unless using expensive 6-axis industrial equipment. The system easily adapts to different sensor shapes and their unique safe operating regions. Most importantly, motion capture enables multiple simultaneous indentations, which would otherwise require multiple robotic manipulators at significantly higher cost and complexity.

We implemented an optical motion capture system with 8 PrimeX 13 cameras operating at 120Hz sampling rate. To maximize accuracy, cameras were positioned 1.5-2m from the sensor rather than wall or ceiling mounting. Verification tests confirmed sufficiently low error rates for tactile sensor calibration.

The entire setup was mounted on an aluminum breadboard with precise leveling to ensure minimal movement during data collection. The tactile sensor was fixed at the center, with remaining area used for calibration markers and reference points. The breadboard's pose was continuously tracked to determine relative position between sensor and indenter.

To simultaneously capture pose and force information, we developed a modular indentation unit with three main components: a replaceable indenter, an ATI Mini45 force/torque sensor, and a marker module (see Fig. \ref{fig:indenters_and_setup}, right). The indenter attaches securely to the force/torque sensor while remaining easily interchangeable for different contact geometries. The marker module features 6-7 retroreflective markers in an asymmetric pattern around the holder, ensuring visibility to mocap cameras even during partial occlusion from manual manipulation.

The indentation module origin was calibrated to the indenter tip, ensuring alignment with simulation values. To prevent data synchronization issues, multiple ATI sensors were connected to a single PC. The entire system was integrated with ROS1, with motion capture operating at 120Hz and ATI force measurements sampled at 200Hz.

During data collection, a human demonstrator holds one or two indentation modules and applies gentle, random contacts to the tactile sensor. These contacts produce varied trajectories while the system records ATI sensor outputs, motion capture poses for both base and indentation modules, and image outputs from the vision-based tactile sensor. All collected data is subsequently analyzed and matched, providing key inputs for FE analysis and generating corresponding image or force/torque outputs to maximize experimental efficiency.

This motion capture-based approach offers significant advantages over traditional robotic or CNC methods, representing natural, varied 6D motions that would be difficult to program with most robotic platforms while enabling fast collection of diverse indentation patterns and multi-contact scenarios.

\subsection{Creating Diverse Indenter Shapes from YCB Dataset}

Vision-based tactile sensors with highly compliant gels experience significant non-linear deformations during contact, presenting a unique challenge not found in traditional, stiffer sensors. While most tactile sensing research uses hard gels with small deformations (typically 1-4mm), our work explores much softer materials that enable larger deformations and potentially richer tactile information. This shift introduces a critical challenge: neural networks trained with limited indenter geometries---such as commonly used spherical indenters---tend to memorize specific deformation patterns rather than learn the underlying physics of large, complex deformations. This overfitting problem severely limits model generalizability to novel object interactions and explains why many sensors avoid very soft gels despite their potential advantages for sensitivity and compliance matching with delicate objects.

To address this fundamental challenge unique to soft gel-based tactile sensors, we developed a methodology to create diverse indenter shapes derived from common objects in the Yale-CMU-Berkeley (YCB) object dataset \cite{calli2017yale}. By utilizing varied contact geometries during training, we force the model to learn more generalizable representations of the relationship between visual deformation patterns and physical properties in large deformation regimes.

\begin{figure}[t]
    \centering
    \includegraphics[width=0.93\linewidth]{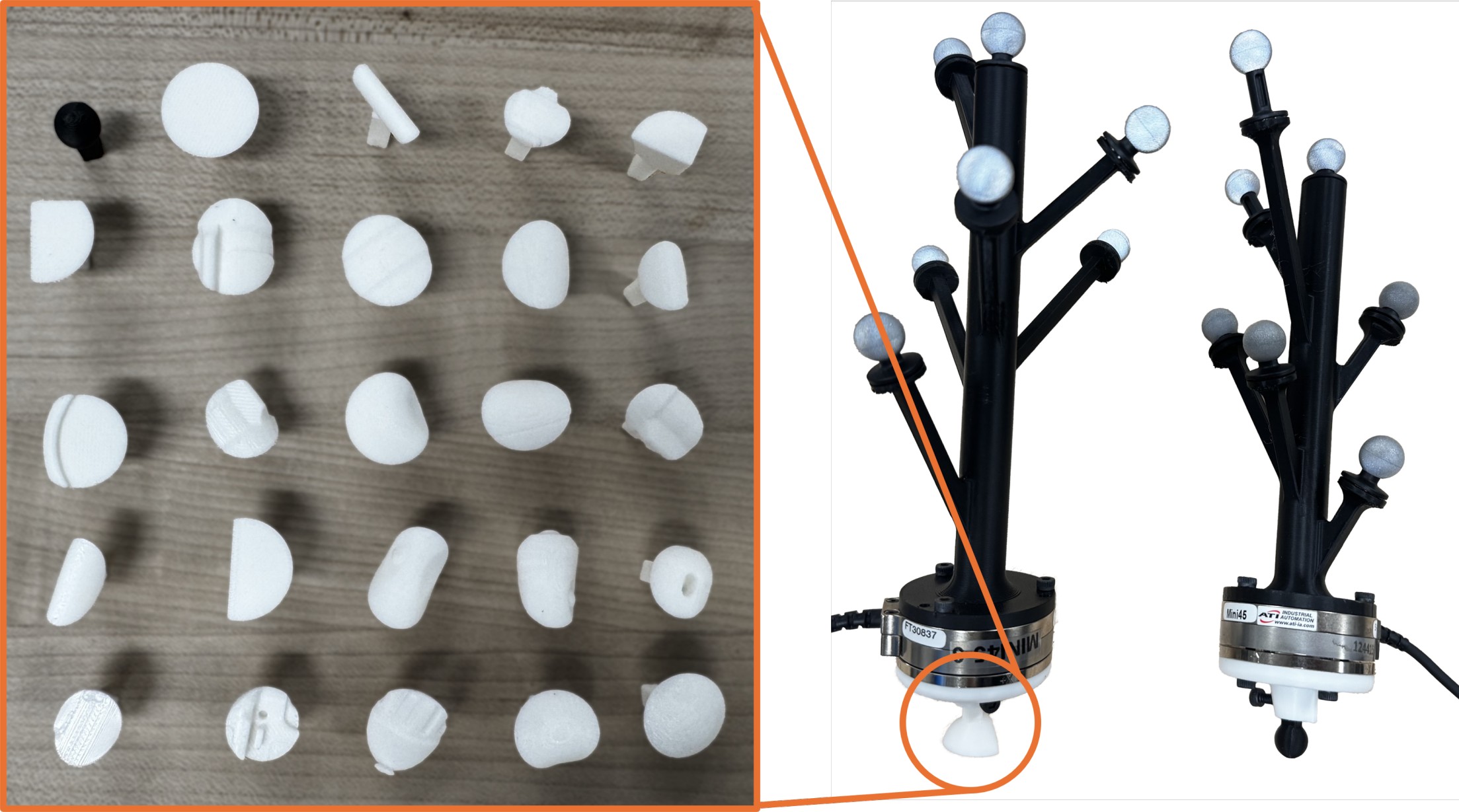}
    \caption{Left: Collection of diverse 3D-printed indenter shapes derived from the YCB dataset, showing various geometric features from sharp edges to smooth curves. Right: Indentation modules used in the motion capture setup, each equipped with an ATI Mini45 force/torque sensor (base), retroreflective markers for pose tracking, and an interchangeable indenter (highlighted by orange circle).}
    \vspace{-1.7em}
    \label{fig:indenters_and_setup}
\end{figure}

Our indenter creation pipeline transforms objects from the YCB dataset into usable indenters for both physical experiments and FE analysis. The process is described by the following integrated approach:
\begin{align}
s_i &= \text{Sphere}(p_i + d \cdot \vec{n}_i, r) \\
I &= \{s_i \cap M \; | \; i = 1, 2, ..., n\}
\end{align}
where $M$ represents the original YCB object mesh, $p_i$ are points sampled from the mesh surface with outward-pointing normals $\vec{n}_i$ that satisfies $\vec{n}_i \cdot (p_i - c_M) > 0$. $\vec{n}_i$ are the outward-pointing normals at those points, $c_M$ is the centroid of the mesh, $d$ is the displacement distance (5mm), $r$ is the sphere radius (15-22mm), and $I$ is the set of all resulting intersection patches. 

To ensure compatibility with both simulation and fabrication, each extracted intersection patch is standardized. The patch is aligned to a standard orientation with its contact normal along the z-axis, scaled by a random factor (350-500) to achieve appropriate dimensions, and positioned so its highest point is exactly 44mm from its base. A cylindrical connector is added to facilitate mounting on the experimental apparatus, and only the largest connected component is retained to ensure topological simplicity. For Abaqus compatibility, the processed meshes are exported in STL format and then converted to STEP format using FreeCAD, which preserves exact geometry. During this conversion, we apply a 90-degree rotation around the X-axis and a 44mm translation to align the indenter correctly for the Abaqus coordinate system. From the original 25 YCB objects, we selected 12 distinct indenter types that provide a wide range of geometric features (Fig. \ref{fig:indenters_and_setup}, left). These were both 3D printed for physical experiments and integrated into our simulation pipeline.

This approach offers key advantages for soft gel tactile sensing. Unlike conventional sensors with stiffer materials where simple indenters may suffice, our compliant gel undergoes complex deformations that require more sophisticated training. By training on diverse indentation patterns, the neural network learns generalizable relationships in the large deformation regime rather than memorizing specific indenter-deformation mappings. The same indenter geometries are used in both physical experiments and simulation, ensuring that sim-to-real discrepancies are not due to shape differences. The indenters are easily interchangeable in our physical setup, allowing for rapid switching between different contact geometries. Most importantly, the selected indenters span a wide range of geometric features—from sharp edges to smooth curves, from flat surfaces to complex contours—challenging the sensor with varied contact conditions that induce the diverse large deformations needed for robust learning.

By systematically varying the indenter shapes used in both simulation and physical experiments, we create a more robust dataset that enables better generalization of our machine learning models to novel contact scenarios, addressing a fundamental limitation that has previously discouraged the use of very soft gels in tactile sensing applications.

% \subsection{Real-to-Sim Correspondence for simulation}

% \edit{Todo: include the keypoint analysis, mocap system implement with ROS (e.g. first collect rosbag data from everything, and then how to find the keypoint out of the force / image threshold / psnr, etc. - maximize the dataset extraction from the running 
% , how to make the input for the abaqus (config file or position match ) , }
% % \subsection{Simulating gel through FE analysis}
% % \edit{Figure: Contact estimation example that matches with real-world simulation \ref{fig:newsensordesign} }

\subsection{Keypoint Extraction for Accurate Motion Simulation}

Establishing accurate correspondence between real-world interactions and FE analysis is critical for developing effective tactile sensor models. We developed a systematic approach to extract representative keypoints from continuous motion capture data, allowing us to faithfully recreate complex indentation sequences in FE analysis. This approach enables random, human-guided indentations to be accurately replicated without requiring precise robot programming.
The keypoint extraction process begins with synchronized data streams from our motion capture system, force/torque sensors, and vision-based tactile sensors. For each indentation sequence, we first detect contact events using a dual-threshold approach. Contact initiation occurs when force magnitude exceeds $F_{\text{contact}} = 1.5\,\text{N},$ and detachment when force drops below $F_{\text{detach}} = 1.1\,\text{N}$.
To capture the complete interaction including the approach phase, we extend each sequence backward by a pre-contact period of 0.3 seconds.

Once contact sequences are identified, we analyze the six-dimensional pose trajectory (position and orientation) to extract meaningful keypoints that characterize the motion. We compute normalized displacements $\Delta\mathbf{p}(t) = \mathbf{p}(t) - \mathbf{p}(t_0)$ and orientation changes $\Delta\mathbf{R}(t) = \mathbf{R}(t_0)^{-1} \cdot \mathbf{R}(t)$, where $\mathbf{p}(t)$ represents the position at time $t$, $\mathbf{R}(t)$ represents the orientation as a rotation matrix, and $t_0$ is the initial time of the sequence. The orientation difference $\Delta\mathbf{R}(t)$ is converted to Euler angles to obtain rotation displacements in a more interpretable format.

To identify keypoints within this normalized trajectory, we apply Butterworth low-pass filtering \cite{butterworth1930theory} to eliminate high-frequency noise while preserving essential motion characteristics. The filter uses cutoff frequency $f_c =f_s/3$ where $f_s$ is the motion capture sampling frequency, with filter order set to 1 to avoid over-smoothing. 

Next, we extract key motion keypoints by finding the prominent peaks in each of the six filtered motion signals (three positional, three rotational), and enforce a minimum spacing (about $1.5 / f_{image}$, where $f_{image}$ is camera frame rate) between peaks so that each detected event is temporally distinct. Keypoints from all six dimensions are combined, with start and end points always included. To prevent redundancy, we enforce minimum time differences between consecutive keypoints and limit the total count when necessary, prioritizing keypoints with larger motion magnitudes.

\begin{figure}[t]
    \centering
    \includegraphics[width=0.92\linewidth]{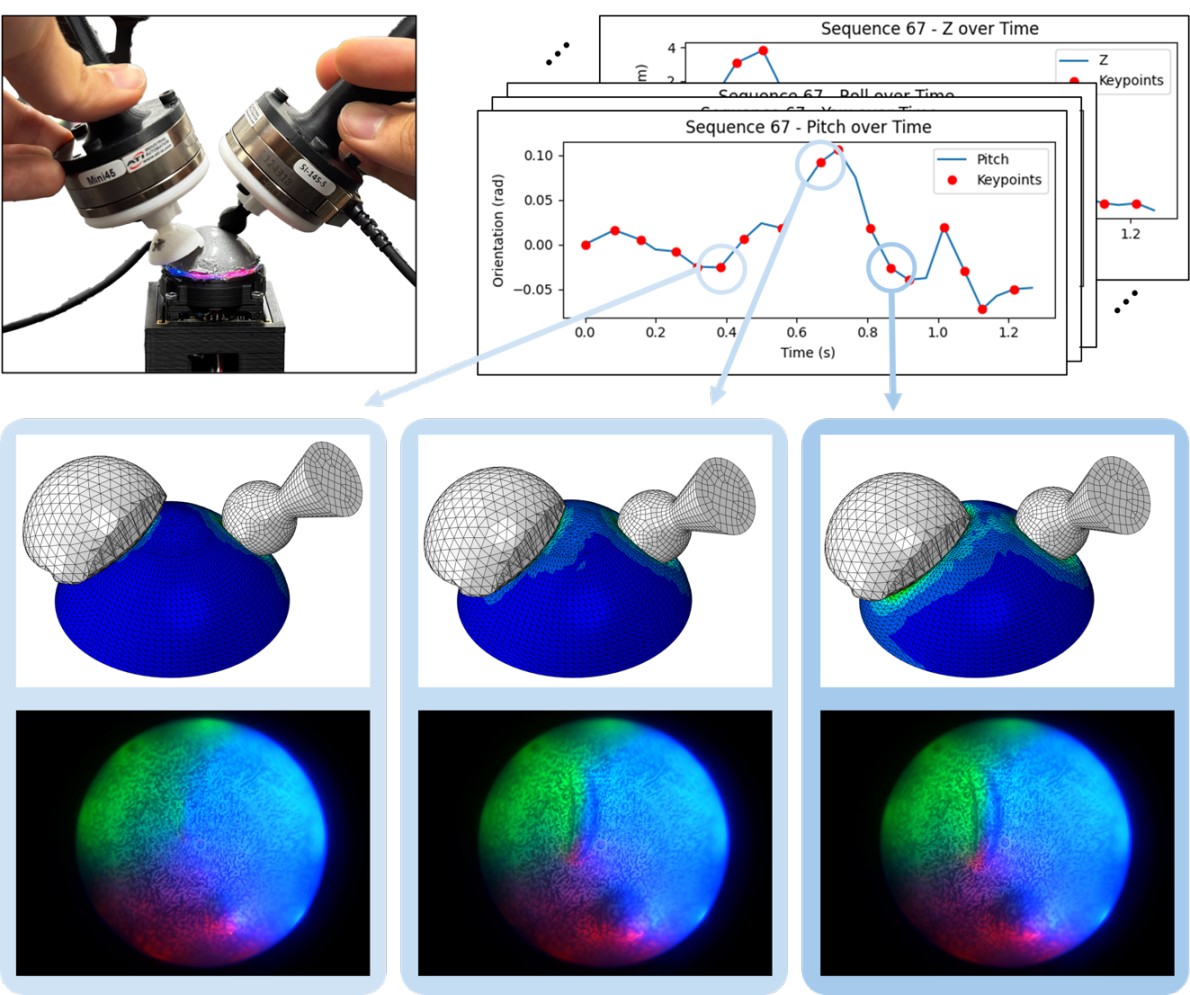}
    \caption{Keypoint extraction process showing the correspondence between motion capture data, FE analysis, and actual sensor readings. The trajectory plots (top right) show extracted keypoints (red dots) from continuous 6D pose data, while the bottom panels demonstrate how these keypoints translate to both FE analysis results (top row) and corresponding tactile sensor images (bottom row) at three different time instances.}
    \vspace{-1.7em}
    \label{fig:sim_keypoint}
\end{figure}

This approach yields a set of keypoints $\{k_1,k_2,\ldots,k_n\}$ that effectively capture the salient moments of the indentation sequence. For each keypoint, we record the timestamp, absolute position and orientation of the indenter, position and orientation relative to the initial pose, corresponding force/torque readings, and the associated tactile sensor image. For handling multiple indenters, we detect contact intervals for each indenter independently and merge these intervals to identify periods of simultaneous contact, enabling accurate modeling of complex multi-contact scenarios. Fig. \ref{fig:sim_keypoint} illustrates this keypoint extraction process, showing how continuous trajectory data is reduced to critical keypoints that preserve essential motion characteristics for both simulation and real sensor correspondence.

The extracted keypoints are crucial for FE analysis for several reasons. They significantly reduce computational requirements by focusing only on critical trajectory points rather than simulating entire dense time series. Keypoints preserve important motion features including direction changes, contact events, and maximum deformations. For multi-step FE analysis, each step transitions between consecutive keypoints, creating efficient yet accurate motion representations.
This keypoint-based approach offers substantial advantages over traditional robotic arm or CNC machine methods for training data creation, representing natural, varied 6D motions that would be difficult to program or execute with most robotic platforms. The system adapts automatically to different gel shapes without requiring complex trajectory planning for shape-specific constraints, enabling fast collection of diverse indentation patterns by leveraging human intuition to explore the interaction space. Multi-contact scenarios using multiple indenters can be easily implemented without requiring multiple synchronized robotic manipulators.

For each extracted keypoint sequence, we generate a configuration file serving as input to the FE analysis. This file contains the initial pose of each indenter relative to the gel coordinate system and the sequence of displacement vectors defining motion between keypoints. The simulation proceeds step-by-step through these keypoints, computing resulting gel deformation, stress distributions, and contact forces at each stage.

The comprehensive pose and force information captured at each keypoint ensures our simulation accurately reproduces physical interactions, including subtle effects like friction, stick-slip phenomena, and complex material deformations occurring during real-world tactile sensing.

\subsection{FE analysis of Hyperelastic Material}

When the surface of the tactile sensor deforms due to an external force from contact with an object, the force is distributed throughout, causing the entire sensor to deform. In this case, measuring the actual deformation of the tactile sensor is extremely difficult due to: (1) occlusion by the external object, (2) the lack of precise measurement tools—such as 3D scanners—that can capture the deformation without accurate segmentation of the scanning results, and (3) the minute magnitude of the deformation. Therefore, the best approach to estimate the deformation of the sensor is to utilize FE analysis.

FE analysis has been developed over decades and is widely used to model the actual deformation of real-world materials at both research and industrial levels. In continuum mechanics, FEA solves equilibrium problems by discretizing a continuum into an assembly of smaller elements and approximating the governing equations over these elements. This approach is based on balancing each element’s internal forces (stresses) with the external forces and the resulting deformations. The linear stress-strain relationships can be expressed using Hooke's law as follows:
\[
\sum_{j=1}^3 \frac{\partial \sigma_{ij}}{\partial x_j} + f_i = 0,\quad \epsilon_{ij} = \frac{1}{2}\Big(u_{i,j} + u_{j,i}\Big),\quad \sigma_{ij} = C_{ijkl}\epsilon_{kl}\,.
\]

The first equation represents the static equilibrium of an infinitesimal element in three dimensions (with \(i,j = 1,2,3\)), ensuring equilibrium in all Cartesian directions. The second equation defines the strain tensor for small deformations using linearization, where \(u_{i,j} = \partial u_i/\partial x_j\) denotes the gradient of the displacement field \(u_i(x)\). The third equation represents Hooke’s law, with \(C_{ijkl}\) being the fourth-order elasticity tensor. For isotropic materials, the elasticity tensor can be expressed via Young’s modulus \(E\) and Poisson’s ratio \(\nu\) as
\[
C_{ijkl} = \frac{E}{2(1+\nu)}\left(\delta_{ik}\delta_{jl} + \delta_{il}\delta_{jk}\right) + \frac{E\nu}{(1+\nu)(1-2\nu)}\,\delta_{ij}\delta_{kl}\,.
\]
This formulation shows that the linear stress–strain relationship is governed by two independent material constants. Assuming a linear stress–strain relationship and small deformations, the FE analysis using the above equations is mathematically convenient. However, this approach is insufficient for describing real-world materials when deformations become large. Materials used in the fabrication of tactile sensors, such as silicone and rubber, can sustain large elastic deformations that exhibit a nonlinear stress–strain response. Even for moderate strains, these materials tend to be much softer at low strains and become stiffer at higher elongations. Such materials, known as hyperelastic or Green elastic materials, require nonlinear material models. In other words, tactile sensors with high deformability require nonlinear material models for accurate FE analysis.

Hyperelastic models account for large deformations in soft materials—such as silicone and rubber-like materials—which exhibit no plasticity. Although a hyperelastic material displays elastic properties such as no energy loss and path-independent unloading, its stiffness and tangent modulus vary with strain; in other words, the material can stiffen or soften as it deforms. Therefore, the stress in a hyperelastic material should be derived from a strain energy density function \(W(I_1, I_2, I_3)\), rather than relying solely on Young’s modulus and Poisson’s ratio, where \(I_i\) denotes the principal invariants of the right Cauchy–Green deformation tensor:
\begin{align}
    I_1 &= \text{tr}(\mathbf{C}) = \lambda_1^2 + \lambda_2^2 + \lambda_3^2\,, \\
    I_2 &= \tfrac{1}{2}\Big[(\text{tr}\,\mathbf{C})^2 - \text{tr}(\mathbf{C}^2)\Big] = \lambda_1^2\lambda_2^2 + \lambda_2^2\lambda_3^2 + \lambda_1^2\lambda_3^2\,, \\
    I_3 &= J^2 = \det(\mathbf{C}) = \lambda_1^2\,\lambda_2^2\,\lambda_3^2 = (\lambda_1\lambda_2\lambda_3)^2\,.
\end{align}
Here, \(I_1\) relates to the change in total stretch (length), \(I_2\) corresponds to area distortions, and \(I_3 = J^2\) is the square of the volume ratio, with \(J = \det \mathbf{F}\) representing the total volume change. For incompressible materials, \(J=1\) (and thus \(I_3=1\)) is fixed. Moreover, \(\mathbf{C} = \mathbf{F}^T \mathbf{F} = \mathbf{U}^2\) is the right Cauchy–Green deformation tensor, where \(\mathbf{F}\) is the deformation gradient with components \(F_{ij} = \partial x_i/\partial X_j\) for the deformed coordinates \(\mathbf{x}\) and the reference coordinates \(\mathbf{X}\). The quantities \(\lambda_i\) are the principal stretch ratios, which are the eigenvalues of the right stretch tensor \(\mathbf{U}\).

By using different combinations of these invariants to represent a strain energy density function \(W\), it is possible to accurately model the hyperelastic properties of materials. The stress tensor is derived from W through the relationship $\sigma = \frac{2}{J}\mathbf{F}\frac{\partial W}{\partial \mathbf{C}}\mathbf{F}^T$, where the derivatives of W with respect to the deformation invariants determine the material's stress response.Several approaches have been developed to express \(W\), either phenomenologically or mechanistically:
\begin{align}
     & \quad \text{Neo-Hookean \cite{treload1975physics}: } W = C_{1}\,(I_1 - 3)\,, \\
    & \quad \text{Mooney-Rivlin \cite{rivlin1948large}: } W = C_{1}(I_1-3) + C_{2}(I_2-3)\,, \\
    & \quad \text{Ogden \cite{ogden1997non}: } W = \sum_{p=1}^{N} \frac{\mu_p}{\alpha_p}\Big( \lambda_1^{\alpha_p} + \lambda_2^{\alpha_p} + \lambda_3^{\alpha_p} - 3 \Big)\,, \\
     & \quad \text{Yeoh \cite{yeoh1993some}: } W = \sum_{i=1}^n C_i (I_1-3)^i + \frac{1}{D_1}(J-1)^2\,.
\end{align}
Here, \(C_i\), \(\mu_p\), \(\alpha_p\), and \(D_1\) are empirical material constants. Among the hyperelastic models, the Yeoh model stands out for its applicability in the FE analysis of silicone rubber in tactile sensors, particularly when dealing with significant deformations. While simplified models like the Neo-Hookean model or linear elasticity, as seen in previous tactile sensing studies \cite{narang2021sim, ma2019dense}, are computationally efficient, they often fail to accurately represent the material behavior of sensors experiencing large strains. In tactile sensing, even seemingly small displacements can lead to large local strains in sensor materials, thereby impacting the accuracy of shape and force reconstruction. The Yeoh model, with its ability to capture the nonlinear stress–strain relationship and strain-hardening behavior of elastomers with a relatively small number of parameters, offers a more robust solution for FE analysis.

Compared to other higher-order models such as the Ogden model, the Yeoh model presents a more practical choice for tactile sensor applications. Although the Ogden model can achieve high accuracy with ample experimental data, its excessive parameterization can lead to overfitting and numerical instability, particularly when applied to complex sensor geometries or limited experimental datasets common in tactile sensing research. The Yeoh model relies only on the first invariant \(I_1\), providing a good balance between accuracy and computational cost. This is particularly valuable in situations where full multi-axial testing is not feasible, such as in the case of human tissue \cite{kaster2011measurement}. Similarly, the Yeoh model’s ability to accurately model elastomer behavior across various strain levels makes it an ideal candidate for silicone elastomeric gels, allowing for more reliable shape estimation and force reconstruction compared to simpler, less accurate models.

\begin{figure}[t]
    \centering
    \includegraphics[width=0.92\linewidth]{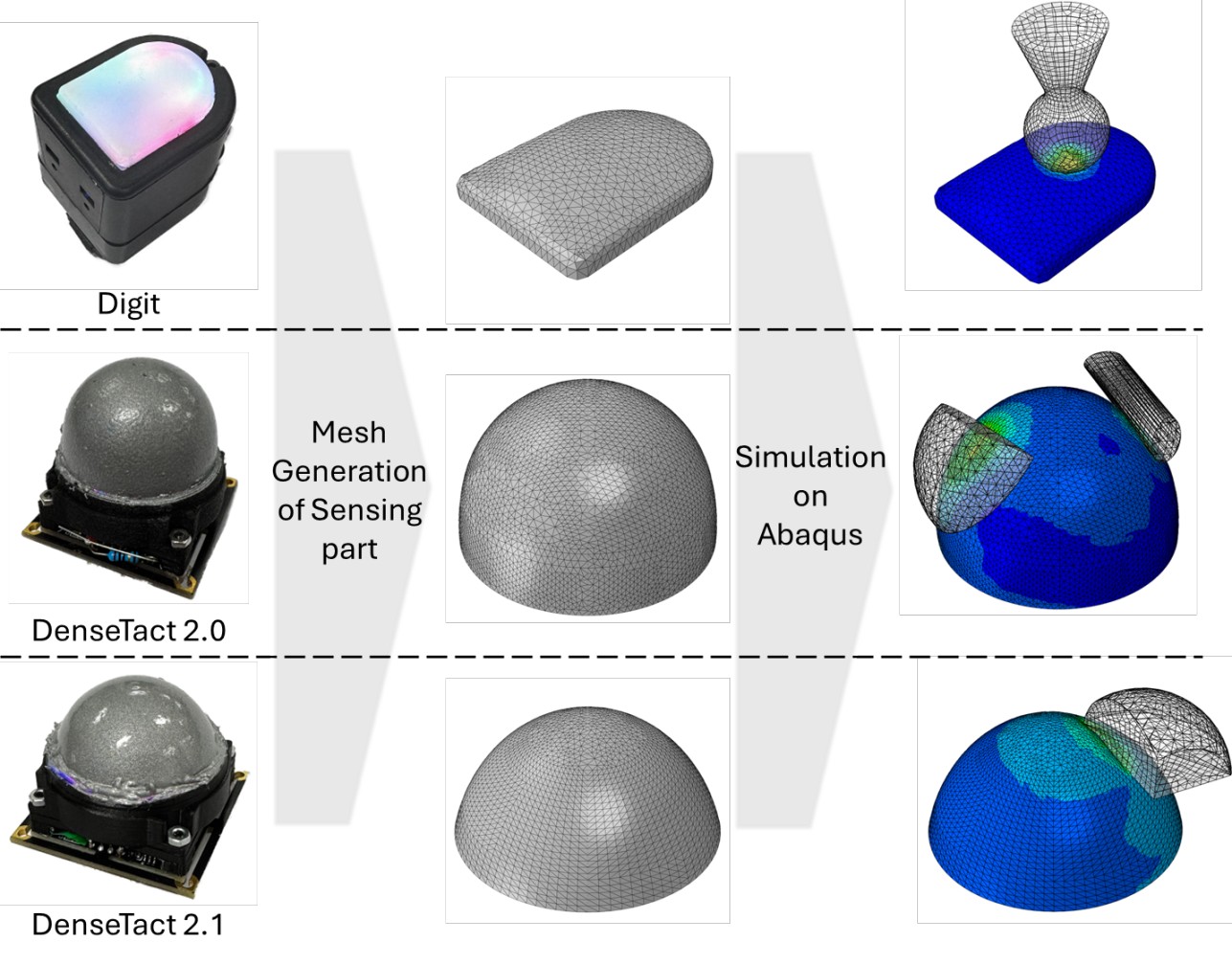}
    \caption{The application of frameworks on various sensors. The real-to-sim framework is easily expandable to various shape of the optical tactile sensors with appropriate boundary condition.}
    \vspace{-1.7em}
    \label{fig:sim_various_sensors}
\end{figure}

\subsection{FE Analysis Simulation Setup}

The FE analysis model was implemented using Abaqus 2024 \cite{Dassault2024AbaqusCAE} to simulate the mechanical response of tactile sensors under various indentation scenarios. 
% This section details the comprehensive modeling approach, including gel formulation, material properties, boundary conditions, mesh strategies, and simulation execution.
The gel component was modeled based on CAD files extracted from each sensor's specifications. These geometries were either determined from available STL files or defined through physical molds of the gel. For sensors with a three-dimensional hemispherical shape, we partitioned the gel volume using strategically placed datum planes to facilitate controlled mesh generation. Specifically, for the hemispherical gel geometries, we created a comprehensive partitioning scheme with horizontal planes at specific heights relative to the gel's total height. Three equally distributed datum planes were created along the vertical axis, with the topmost plane adjusted to approximately two-thirds of the maximum height to avoid creating excessively thin sections near the apex. Additionally, we created angular partitions by generating rotational datum planes at 45-degree intervals around the vertical axis, starting from an initial YZ reference plane. This approach allowed for systematic cell partitioning that accommodated the curved geometry.

For computational efficiency, we simplified the gel model to a single homogeneous silicone material, omitting the thin reflective surface layer typically present in vision-based tactile sensors. This simplification is justified as these reflective layers and the applied pattern underneath are extremely thin compared to the bulk gel material and have minimal impact on the overall mechanical response. The reflective layer serves to block external light and enhance contrast for the vision system, but its mechanical contribution to deformation is negligible.

The gel material was modeled using the Yeoh hyperelastic constitutive model, which effectively captures the nonlinear elastic behavior of silicone rubber under large deformations. Material parameters were either directly specified in the configuration file or derived from experimental uniaxial and biaxial test data when available. The material density was set to $1.07\times10^{-9}$ (in consistent units), representing the typical density of silicone elastomers used in tactile sensors.

The friction coefficient between the indenter and gel surface was set to 2.2 based on findings from \cite{lee2023influence}, who investigated silicone rubber friction characteristics. This relatively high friction value accurately represents the sticky nature of silicone surfaces. The contact interaction was defined using a penalty formulation with normal ``hard'' contact that allows separation after contact and prevents penetration during compression.

\begin{figure}[t]
    \centering
    \includegraphics[width=0.95\linewidth]{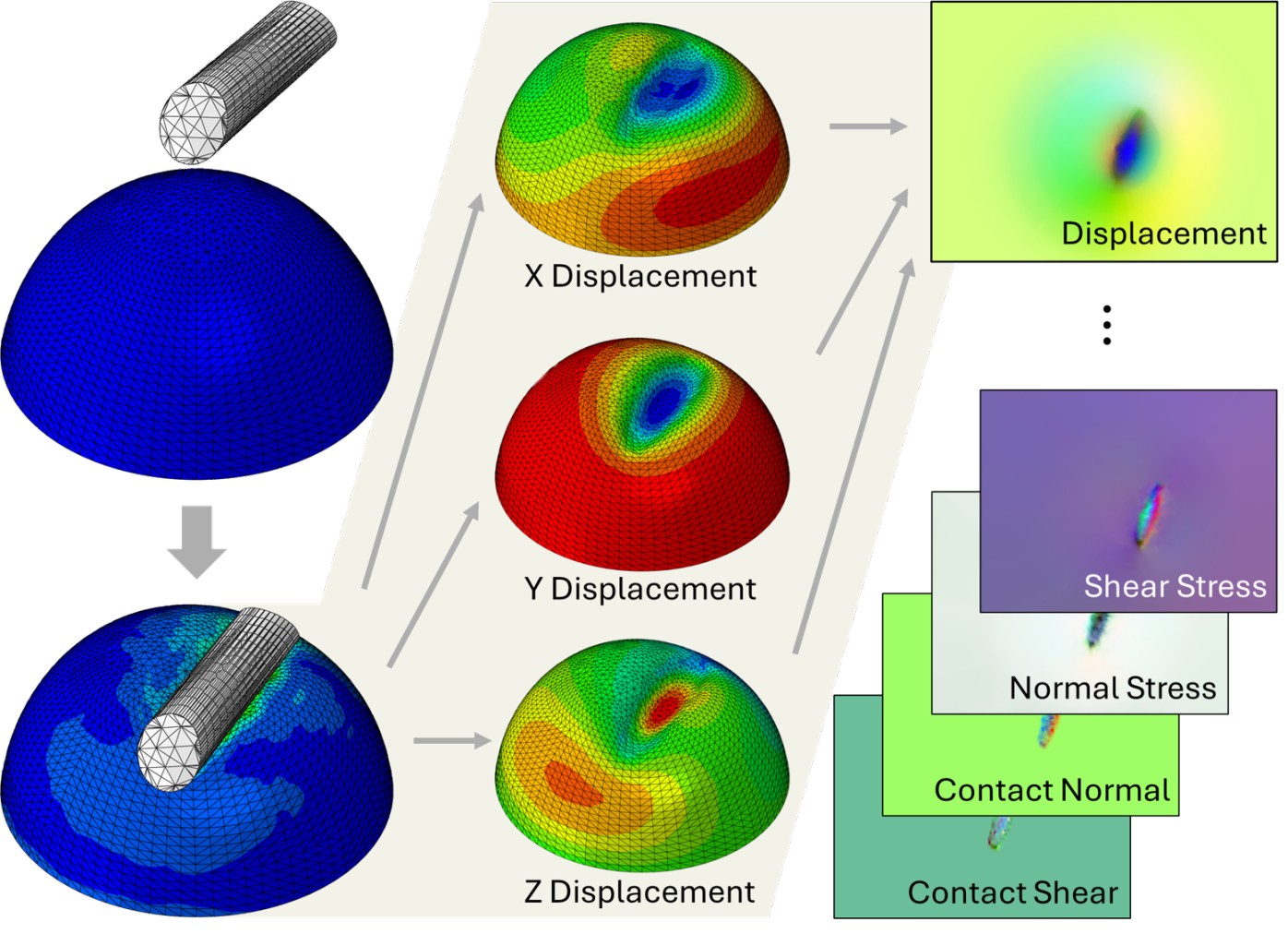}
    \caption{Conversion of 3D FE analysis results to 2D tactile image representations. The left side shows 3D displacement fields (X, Y, Z components) from the deformed gel simulation, while the right side displays the corresponding 2D image encodings for displacement, normal stress, shear stress, contact normal forces, and contact shear forces used for neural network training.}
    \vspace{-1em}
    \label{fig:sim_abaqus2img}
\end{figure}

The simulation was configured to extract comprehensive mechanical response data, including stress tensors, strain components, displacements, reaction forces, and detailed contact information. These output variables provide a complete picture of the gel's deformation behavior and the resulting tactile response. Specifically, we requested field outputs for stress (S), strain (E), displacement (U), reaction forces (RF), rotation (UR), and numerous contact variables (CNAREA, CNORMF, CSHEAR, CPRESS, CSTATUS) to characterize the contact interface completely.

For element selection, the indenter was modeled as a discrete rigid surface (R3D3 and R3D4 elements) since the 3D-printed PLA material used for physical indenters is significantly stiffer than the silicone gel, resulting in negligible deformation of the indenter compared to the gel. The gel was modeled using C3D8H hybrid elements for hexahedral regions, with C3D6 (6-node wedge) and C3D4 (4-node tetrahedral) elements used where geometrically necessary. The hybrid formulation is particularly suitable for nearly incompressible hyperelastic materials like silicone.

Boundary conditions were carefully implemented to accurately represent the physical setup. The bottom surface of the gel was fully constrained with an encastre boundary condition (all translational and rotational degrees of freedom fixed), reflecting the typical mounting of tactile sensors on rigid supports. This was implemented by selecting nodes within a small tolerance (0.1 mm) of the XZ plane and applying fixed constraints, ensuring proper anchoring without over-constraining the model.

The assembly and simulation steps were configured based on extracted pose keypoint data. For each indenter, the reference position and orientation were established through a sequence of translations and rotations defined by the configuration data. The indenter motion was controlled through a series of displacement boundary conditions applied to the reference point of each rigid indenter. These displacement vectors were derived from motion capture or tracking data obtained from physical experiments, ensuring that the simulation accurately reproduces real-world indentation sequences.

Our mesh generation strategy was optimized to balance computational efficiency with solution accuracy. We implemented a non-uniform mesh density distribution with finer elements (approximately 0.7 mm characteristic length) near the gel surface where contact occurs and more detailed stress-strain information is needed. This gradually transitioned to coarser elements (approximately 2.0 mm) in the interior regions where stress gradients are less severe. This approach significantly reduced computational demands without compromising the accuracy of tactile information at the sensor surface. The transition between mesh densities was controlled using a size growth rate of 1.97, allowing for smooth transitions between regions of different mesh density.

A key aspect of our implementation is the capability to simulate multiple indenters simultaneously, which is essential for realistic tactile sensing during dexterous manipulation. Most real-world manipulation scenarios involve contact with multiple surfaces or objects, and our model seamlessly accommodates this complexity. Each indenter can follow independent trajectories with six degrees of freedom (three translational and three rotational), enabling the simulation of complex interaction scenarios such as multi-finger grasping or manipulation involving multiple contact points. The displacement sequences for each indenter are managed through separate boundary condition specifications that can be updated independently at each simulation step.

The simulations were executed on a high-performance computing setup utilizing 20 CPU cores and 2 RTX 4090 GPUs, which enabled efficient processing of the computationally intensive nonlinear analyses. Each simulation step was carefully monitored through Abaqus message files to ensure proper convergence and solution quality. Upon completion of the simulations, comprehensive data extraction was performed to capture the complete state of the model at each time increment. This includes exporting deformed meshes as STL files and detailed nodal and elemental results as CSV files, facilitating further analysis and visualization.

The frame-by-frame data extraction approach enables precise temporal correlation between simulation results and real-world experimental data, which is crucial for validation and for developing accurate mappings between physical sensor responses and simulated mechanical states. This comprehensive data set forms the foundation for advanced tactile sensing algorithms and enables physics-based interpretation of sensor outputs during complex manipulation tasks.

\subsection{Data Extraction with Dynamic Baseline Image Adaptation}

Accurate tactile sensing requires establishing precise correspondences between the deformed gel image, the undeformed reference state, and the 3D physical world. We developed systematic methods to address two critical aspects of this correspondence: dynamically maintaining an appropriate baseline (undeformed) image despite hysteresis effects, and mapping between 2D image coordinates and 3D gel surface coordinates.

% \subsubsection{Dynamic Baseline Image Adaptation}

Vision-based tactile sensors detect deformation by comparing the current (deformed) image with a reference baseline image representing the undeformed state. However, silicone gels exhibit hysteresis---they do not immediately return to their original shape after deformation but rather recover gradually over time. This creates a fundamental challenge: using a single static baseline image would introduce increasing errors as the gel's resting state evolves throughout repeated interactions.

To address this challenge, the baseline of undeformed image is dynamically adapted to continuously monitor the gel's state and update the reference image when appropriate. We first establish a master baseline by capturing the gel in its initial undeformed state. For robustness, we capture multiple frames and select the most stable representation by comparing consecutive images with $W$ and $H$ as width and height of the images:
\begin{align}
\text{MSE}(I_1, I_2) = \frac{1}{WH}\sum_{x=0}^{W-1}\sum_{y=0}^{H-1}\Bigl(I_1(x,y) - I_2(x,y)\Bigr)^2,
\end{align}

Next, We continuously evaluate the gel's state using MSE between the current image and master baseline, plus force magnitude from the force/torque sensor. Based on these metrics, we classify three states: contact (high MSE, force above threshold), detachment (force below threshold but gel may not have recovered), and stable (low MSE, suitable for baseline update).
During the detachment state, we monitor image stability. If MSE remains below a threshold ($\text{MSE}_{\text{threshold}} = 8$)or sufficient time, we update the current baseline to this new stable state.

This adaptive approach ensures that even after interactions causing temporary or permanent deformations, the system maintains an accurate reference for detecting new contacts—essential for reliable force and deformation estimation during extended manipulation tasks.

\subsection{2D-3D Correspondence Mapping}

Another critical challenge is establishing the mapping between 2D image coordinates and 3D physical coordinates on the gel surface. This mapping is essential for precisely localizing contacts and accurately estimating forces and deformations.

We developed a systematic calibration procedure that establishes this mapping through controlled physical measurements. First, we identify the gel center in the image plane by analyzing motion capture data from the experimental setup, identifying the highest point of the gel, and projecting it onto the image plane. We refine this center point through visual validation and interactive adjustment to account for any minor misalignments between the camera and gel.

Next, we collect correspondence points by gently poking the gel surface at various locations using a calibrated probe. For each poke, the 3D position of the contact point is recorded using the motion capture system, the corresponding 2D location in the tactile sensor image is identified by detecting local deformation, and multiple pokes are performed to ensure uniform coverage of the gel surface. 

From these correspondence points, we establish a mapping between spherical coordinates in 3D space and image coordinates. We represent the 3D surface using two angular coordinates: $\theta$, the angle from the vertical axis (related to the radial distance in the image), and $\phi$, the azimuthal angle around the vertical axis (related to the angular position in the image).
\begin{figure}[t]
    \centering
    \includegraphics[width=0.94\linewidth]{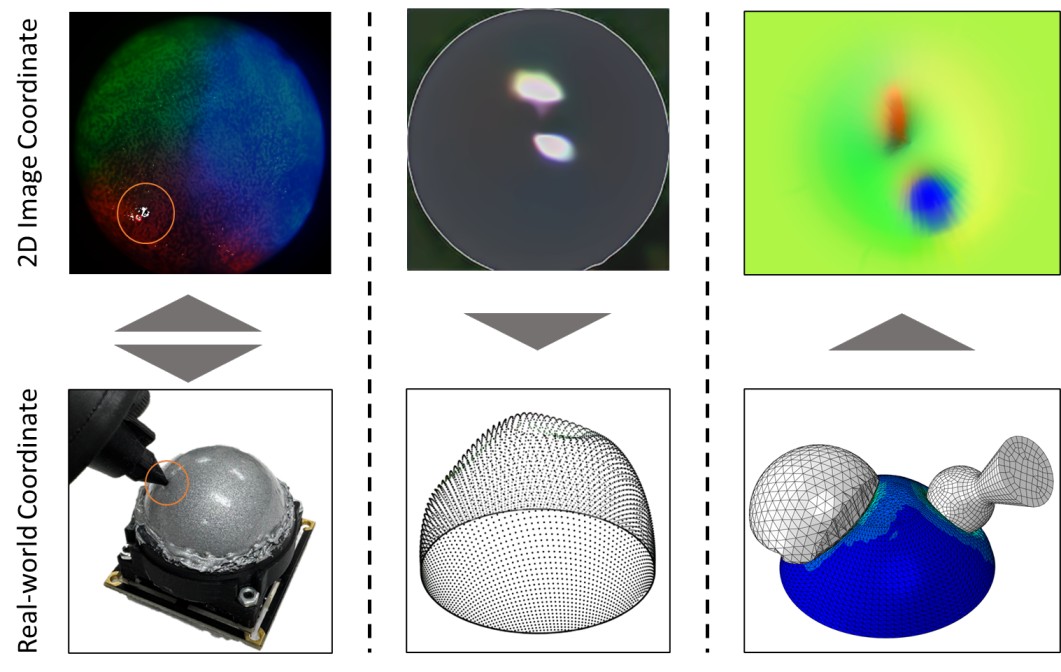}
    \caption{2D-3D correspondence mapping between image coordinates and physical contact points on the gel surface. }
    \vspace{-1.7em}
    \label{fig:2d3d}
\end{figure}
We use Gaussian Process Regression (GPR) to learn the relationship between image coordinates and these spherical coordinates:
\begin{align}
\text{dist} &= \sqrt{(x - x_c)^2 + (y - y_c)^2}, \\
\theta &= f_{\text{GPR}}(\text{dist}), \\
\phi &= \arctan2(x - x_c, y - y_c),
\end{align}
where $(x_c, y_c)$ is the center of the gel in the image plane. This approach accounts for the non-linear relationship between image distance and the corresponding angle on the gel surface.

The resulting mappings are stored as dense lookup tables, $\theta_{\text{mask}}$ and $\phi_{\text{mask}}$, which provide the corresponding spherical coordinates for each pixel in the image.

The 2D-3D correspondence mappings are illustrated in Fig. \ref{fig:2d3d} and Fig. \ref{fig:sim_abaqus2img}. The first row of Fig. \ref{fig:2d3d} demonstrates the correspondence between image and real-world contact points. The second row shows how, based on GP regression, predicted stress tensor images can be mapped into pointclouds where each point contains stress tensor values encoded as RGB values. The final column illustrates the correspondence from simulation to training dataset as 2D images. This correspondence enables precise localization of contacts in 3D space from 2D image features, supports accurate geometric interpretation of visual deformations, facilitates registration between simulation outputs and actual sensor images, and allows transformation between contact frames and world coordinates.

The 2D-3D correspondence ensures that our neural network models can effectively learn the relationship between visual patterns and physical quantities like force and displacement. By addressing both the temporal evolution of the baseline state and the spatial mapping between image and physical coordinates, our approach provides a robust foundation for accurate tactile sensing across diverse interaction scenarios.

\subsection{Force and Deformation Field Mapping}

Establishing accurate correspondence between 3D simulation outputs and 2D tactile images requires a sophisticated approach to data representation and transformation. Our method converts the dense 3D FE analysis results from Abaqus into spatial force and displacement fields that can be directly compared with tactile sensor images.
The transformation process begins by applying our previously established 2D-3D correspondence maps ($\theta_{\text{mask}}$ and $\phi_{\text{mask}}$) to convert between image coordinates and spherical coordinates on the gel surface. For each simulation output frame, we extract comprehensive mechanical data including stress tensors ($\sigma_{xx},\;  \sigma_{yy},\;  \sigma_{zz}, \; \sigma_{xy},\;  \sigma_{xz}, \; \sigma_{yz}$), displacement vectors ($d_x,\;  d_y,\; d_z$), and contact information (CNAREA, contact normal forces $(F_{n,x},\; F_{n,y},\; F_{n,z})$, and contact shear forces $(F_{s,x},\; F_{s,y},\; F_{s,z})$).
% The transformation process begins by applying our previously established 2D-3D correspondence maps ($\theta_{\text{mask}}$ and $\phi_{\text{mask}}$) to convert between image coordinates and spherical coordinates on the gel surface. For each simulation output frame, we extract comprehensive mechanical data including stress tensors (S-S11, S-S22, S-S33, S-S12, S-S13, S-S23), displacement vectors (U-U1, U-U2, U-U3), and contact information (CNAREA, contact normal forces, contact shear forces).

To create accurate mapping between simulation nodes and the tactile sensor image, we employ a multi-stage process:
First, we build a spatial k-d tree \cite{maneewongvatana1999analysis} from the 3D positions of all simulation nodes. For each vertex in the gel's surface mesh, we find the nearest simulation node using this tree and transfer the mechanical properties from the simulation data to the mesh vertex. This creates a dense sampling of mechanical information across the entire gel surface.

Using our previously established 2D-3D correspondence maps, we create a regular grid defined by the phi and theta coordinates that represents the 2D layout matching the tactile sensor's view. We then use linear interpolation methods to map the scattered simulation data onto this grid. To handle the circular nature of the phi coordinate, we extend our data by replicating points with adjusted phi values, ensuring smooth and continuous interpolation even at boundaries.

% Next, we compute the spherical coordinates ($\phi, \theta$) for each mesh vertex based on its 3D position and adjust $\phi_{\text{mesh}}$ to ensure the proper range:
% \begin{align}
% r_{\text{real}} &= \sqrt{x_v^2 + z_v^2}\\
% \theta_{\text{mesh}} &= \arctan2(y_v, r_{\text{real}})\\
% \phi_{\text{mesh}} &= \arctan2(x_v, -z_v) \end{align}

% Instead of using complex equations, we create a regular grid defined by the phi and theta coordinates (which we obtain from our correspondence masks). This grid represents the 2D layout that matches the tactile sensor’s view. We then use a standard interpolation method (such as griddata) to map the scattered simulation data-such as stress and displacement values-onto this grid. To handle the circular nature of the phi coordinate (where values wrap from 2$\pi$ back to 0 via $\phi_{\text{mesh}} = (\phi_{\text{mesh}} + 2\pi) \mod 2\pi$), we extend our data by replicating the points with adjusted phi values. This extension ensures that the interpolation is smooth and continuous, even at the boundary.

This ensures that interpolation near the boundary correctly considers points from both sides of the discontinuity.
From these interpolated fields, we create five distinct types of force/deformation visualizations:
\begin{itemize}
    \item Normal stress images: RGB channels represent normal stress components $\sigma_{xx},\;  \sigma_{yy},\;  \sigma_{zz}$
    \item Shear stress images: RGB channels represent shear stress components $\sigma_{xy},\;  \sigma_{xz},\;  \sigma_{yz}$
    \item Displacement images: RGB channels represent displacement components $d_x, \;d_y, \;d_z $
    \item Contact normal force images: RGB channels represent contact normal force components
    \item Contact shear force images: RGB channels represent contact shear force components
\end{itemize}
Fig. \ref{fig:sim_abaqus2img} illustrates this transformation process, showing how the 3D displacement fields from FE analysis are converted into the corresponding 2D image representations used for training our neural network models.

For each field type, we apply Gaussian smoothing ($\sigma_{stddev} = 2$) to reduce noise and enhance visual coherence. We normalize each channel using both global statistics (derived from the entire dataset) and local statistics (specific to each image). The global normalization ensures consistent interpretation across the dataset, while local normalization enhances contrast for individual images.

This method transforms detailed 3D simulation results into clear 2D images. By matching every mesh vertex to a specific location on a 2D grid that mirrors the tactile sensor’s camera view, it preserves the contact information from 3D space. 2D image representation enables direct comparisons between simulated data and real tactile measurements, while also maintaining consistent spatial relationships and mechanical information across the dataset.

For multi-indenter scenarios, we track which gel surface nodes are in contact with each indenter. When multiple indenters make contact simultaneously, we apply k-means clustering to group contact points and assign each cluster to the nearest indenter. This allows us to correctly attribute contact forces to the appropriate indenter, ensuring accurate force reconstruction even during complex interaction scenarios.

The final step involves storing the mapping bounds for each image type in a JSON file, which records the minimum and maximum values used for normalization. These bounds are essential for consistent interpretation during model training and inference, allowing the neural network to properly understand the relationship between pixel values and physical quantities.

\subsection{Bridging Sim-to-Real Using Sensor Measurements}

To bridge the gap between simulation and reality, we developed a systematic force calibration method that leverages the ATI force/torque sensor measurements to refine the simulation outputs. This calibration is critical for achieving accurate force estimation in practical applications, as it accounts for differences between idealized simulation conditions and real-world physical interactions.
Our calibration approach addresses several key challenges in tactile sensing: the coordinate frame mismatch between the ATI sensor and the global reference frame, variations in contact mechanics between simulation and reality, and the distribution of forces across multiple contact points. The resulting calibrated data provides a more reliable ground truth for training neural networks.
The calibration process begins by transforming force measurements from the ATI sensor's coordinate frame to the global reference frame using $\mathbf{F}_{\text{global}} = \mathbf{R}_{\text{ati\_to\_global}} \cdot \mathbf{F}_{\text{ati}}$, where $\mathbf{R}_{\text{ati\_to\_global}}$ is the composite rotation matrix combining the ATI-to-indenter and indenter-to-global transformations. This transformation accounts for the orientation of both the sensor mounting and the indenter itself.

\begin{figure*}
    \centering
    \includegraphics[width=0.92\textwidth]{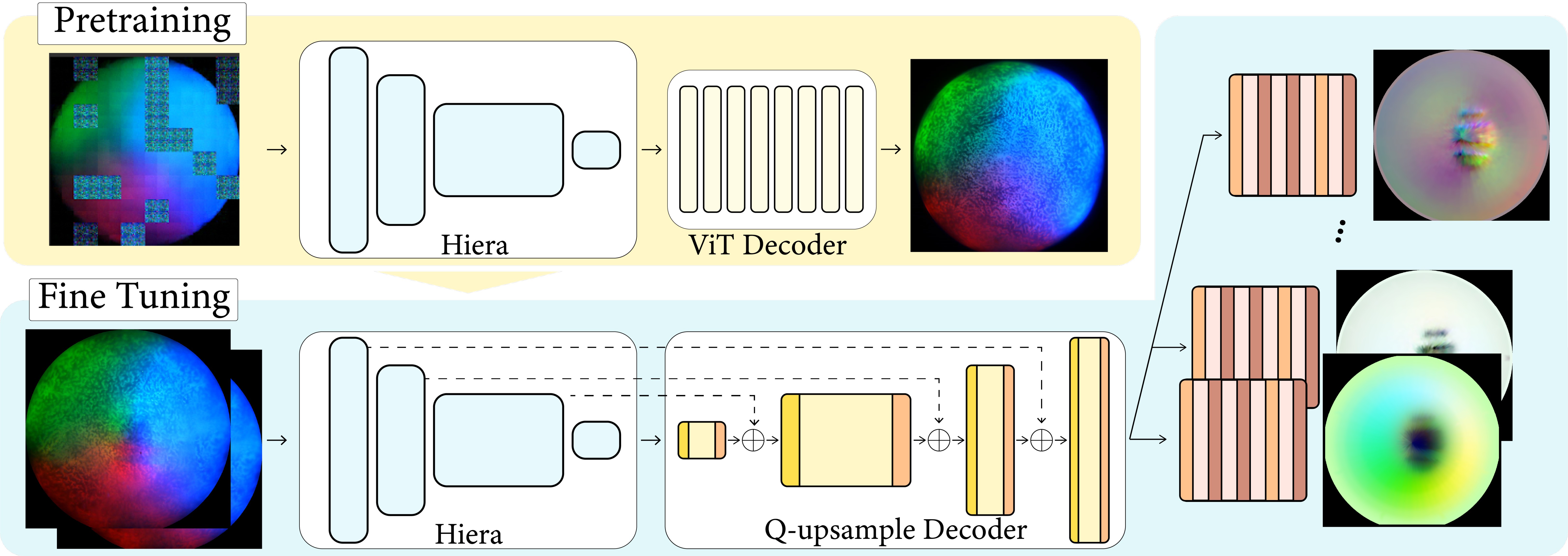}
    \caption{The top row illustrates our self-supervised pretraining: masked or partial tactile images are passed through the Hiera encoder and a ViT-style decoder to reconstruct the full sensor observation. The bottom row shows the fine-tuning stage, where the pretrained Hiera backbone is paired with the proposed Q-Upsample decoder and separate lightweight convolutional heads to produce dense per-pixel predictions of displacement, contact forces, and stress tensors.}
    \vspace{-1.7em}
    \label{fig:network}
\end{figure*}  

Once we have the global-frame forces from the ATI sensor, we compare them with the reaction forces computed by Abaqus and compute component-wise offset ratios. For each active indenter, we extract the simulation reaction forces (RF-RF1, RF-RF2, RF-RF3) and calculate $\text{r}_i = \frac{-\mathbf{F}_{\text{sensor},i}}{\mathbf{F}_{\text{abaqus},i}}$, where the negative sign accounts for the convention difference between sensor measurements and simulation reaction forces. These offset ratios are then directly applied to correct the contact force components. These offset ratios are then applied to correct the contact force components:
\begin{align}
\mathbf{F}_{\text{normal},i} &= \mathbf{F}_{\text{abaqus, normal},i} \cdot r_i \\
\mathbf{F}_{\text{shear},i} &= \mathbf{F}_{\text{abaqus, shear},i} \cdot r_i
\end{align}

Where $\mathbf{F}_{\text{normal},i}$ and $\mathbf{F}_{\text{shear},i}$ are corrected contact normal force and contact shear force. 

For multi-indenter scenarios, the calibration becomes more complex as we need to identify which contact points correspond to each indenter. We solve this by clustering the contact points based on their spatial distribution and assigning each cluster to the nearest indenter. The appropriate offset ratio is then applied to each contact point based on its assigned indenter.

This calibration approach accounts for systematic differences between simulation and reality such as material property variations and friction discrepancies, ensures consistency between external sensor measurements and tactile-based estimates, and provides more accurate ground truth for neural network training.

To bridge the sim-to-real gap that inevitably exists between simulation-based predictions and real-world measurements, we implement a force correction methodology that requires minimal calibration points. The correction pipeline captures synchronized pairs of model-predicted forces and ground-truth measurements across various contact locations, along with corresponding 2D contact positions in the image plane. A random forest regressor with zero-constraint enforcement learns the mapping between those paired measurements with consideration of both the magnitude and position-dependent aspects of the force difference. This approach effectively compensates for systematic errors from material property variations, minor differences in sensor construction, and sensor aging effects. This calibration procedure can be performed with as few as 5-7 measurement points and stored for ongoing use, providing robust correction that maintains accuracy even as environmental conditions change. 

\subsection{Learning Sensor Model}

We train a model $Y = f_\theta(X) \in \mathbb{R}^{H \times W \times C}$ to jointly predict dense output for displacement, contact force, and stress tensors, as shown in Fig. \ref{fig:network}.

% Maybe a figure here would be helpful?
Contrary to prior works that leverage large vision transformers (ViTs) \cite{higuera_sparsh_2024, sharma2025self}, we choose the smaller hierarchical vision transformer Hiera~\cite{Hiera} as the encoder. Hiera removes components of existing hierarchical ViTs that slow it down and hinder its performance compared to vanilla ViTs and CNNs, and creates a lightweight encoder that consistently outperforms standard ViTs. Hiera is an attractive encoder for dense tensor reconstruction, as it naturally learns features at multiple scales, allowing for learning of low level features at a high spatial resolution, and high level features deeper in the network. Paired with an expressive decoder, this suggests Hiera to be useful for reconstructing full deformation and stress tensor without losing information. 

Decoding information from the multi-level features of Hiera requires leveraging the features without loss of information. Prior work, such as the DPT head \cite{DPT} upsamples features from the coarse level and fuses with fine level predictions. The DPT head requires a vanilla ViT, which needs extra operations to preserve the spatial proximity of ViT patches. However, because Hiera preseves the 2D structure in each stage, we propose a \textit{Q-Upsampling module} to naturally decode multi-level Hiera features.

Concretely, we take spatial features from Hiera and feed them through an attention block, which ensures that relevant spatial regions of a tactile image are properly focused on at different levels. We then fuse the output features with features of the same dimension from the encoder. This enables for retaining of fine-grain details deeper into the decoder. Altogether, the proposed encoder-decoder structure requires minimal operations done to the encoder features, similar to traditional CNN-based U-Nets \cite{ronneberger2015u}. At the end of a network, we construct multiple convolution heads for each modality, including displacement, normal force, shear force, shear stress, and normal stress, leading to a $15$-dimensional spatial output.

% \subsubsection{simulation to the real world result correspondence}

% \subsubsection{data augmentation for pretraining}

% % Pre-train stage ; Maybe it is better to put this into implementation details.
% We use MAE~\cite{MAE} to pre-train the encoder part. During the pre-training stage, we use a ViT decoder to recover the missing pieces. In the fine-tuning stage, this decoder is replaced by the Q-Upsample decoder.

% blender side - two sensor 

% \subsection{Calibration estimating method}

% \subsubsection{general structure}
% input undef img, def img, how to select undef img and deformed image, 
% output correspondence,
% ati sensor matching with abaqus result (for contact normal force and shear force)

% \subsubsection{network structure}
% encoder / decoder structure selection reason, hiera and densenet, 

% \subsubsection{pretraining}
% - (Matt) mention the benefits of using vision transformer
% - (Matt) why use Hiera instead of vision transformer?

\subsection{Contact Clustering and Force Distribution Analysis}

To effectively detect and characterize multi-contact interactions with objects of varying properties, we developed a clustering methodology based on the force field outputs from the calibrated sensor. From the contact forces information across the sensor surface, the distinct contact regions are kept tracked. 

From preset force magnitude threshold (typically 0.005N), we first apply a binary contact mask based on the force magnitude from contact shear and normal force. Once the contact mask is generated, we apply connected component analysis using an 8-connectivity structure to identify distinct contact regions. For each identified cluster, we compute centroid, size, total force, and magnitude of the cluster. 

To enhance robustness for sensor-based control, we incorporate additional geometric and force-based metrics for each cluster, including perimeter length, and circularity.  % ($4\pi \cdot area / perimeter^2$)
 For using the cluster info for control, we prioritize clusters based on both force magnitude and spatial consistency.

\subsection{Analysis of Contact Workspace and Optimal Fingertip Pose Selection}

We analyze the contact workspace for two-finger control to determine optimal joint configurations that maximize dexterity while ensuring reliable tactile sensor integration. This analysis is critical for integrating calibrated sensor outputs into precise manipulation control schemes.
Joint configurations are sampled within their respective limits, and forward kinematics computes fingertip positions and orientations. The manipulability measure, metrics introduced by \cite{yoshikawa1985manipulability}, is evaluated as $w = \sqrt{\det\left(J J^T\right)}$, where high $w$ values indicate dexterous configurations responsive to joint variations.

Collision detection using the hpp-fcl library \cite{coalweb} approximates each fingertip as a sphere with 10mm radius and contact is assumed when the collision distance satisfies $d_{\text{FCL}} \leq 1mm$. The contact point and normal are computed as:
\[
p_c = \frac{1}{2}\left(p_{\text{thumb}} + p_{\text{index}}\right), \quad
\mathbf{n} = \frac{p_{\text{thumb}} - p_{\text{index}}}{\| p_{\text{thumb}} - p_{\text{index}} \|}.
\]

Configurations are filtered by distance ($d \leq 0.06$ m for effective pinching) and fingertip alignment computed via $\text{alignment} = \mathbf{z}_{\text{thumb}} \cdot \mathbf{z}_{\text{index}}$, where near-zero or negative values indicate favorable pinching conditions.

For tactile sensor integration, we adopt the strategy of determining optimal sensor pose on existing manipulators rather than designing custom sensors, allowing compatibility across multiple hand designs. We identify thumb poses (pitch angle and vertical offset) that maximize both manipulability and tactile sensor effectiveness.

The optimization process systematically sweeps through candidate thumb poses by modifying the URDF with discrete increments of pitch and $z$-offset. Each variant is evaluated by counting stable thumb-index contacts (via collision checks) and computing manipulability scores. We seek configurations providing high mean manipulability and large sets of contact points aligned along a principal axis suitable for grasping.

Principal axis computation uses RANSAC-like cylinder fitting to identify the 3D line maximizing contact points within a 1 cm radius. This ``best inlier axis'' represents the most reliable thumb-index opposition direction. The number of inlier points serves as a proxy for robust contact establishment across diverse configurations.

The final selection balances two metrics: mean manipulability (reflecting thumb dexterity) and total principal inliers (indicating consistent contact reliability). As shown in Figure \ref{fig:manipulability_vs_z_pitch}, the configuration with $z=4.3$ cm and pitch $=0.0^{\circ}$ yields the largest set of well-aligned contact points while maintaining competitive manipulability.

This thumb pose selection is critical for robust sensing and dexterity. The embedded tactile sensor must maintain stable contact across various grasps while maximizing manipulability for effective maneuvering. Few existing hand designs optimize vision-based fingertip sensor placement for both sensing and kinematic performance, often resulting in suboptimal usage. Our systematic evaluation of contact configurations and manipulability demonstrates improved tactile data quality and overall hand dexterity, enabling subsequent tasks like selective string grasping and force difference detection.

\section{Result }

\begin{figure}[htb]
% \vspace{-1em}
\centering
\includegraphics[width=0.9\linewidth]{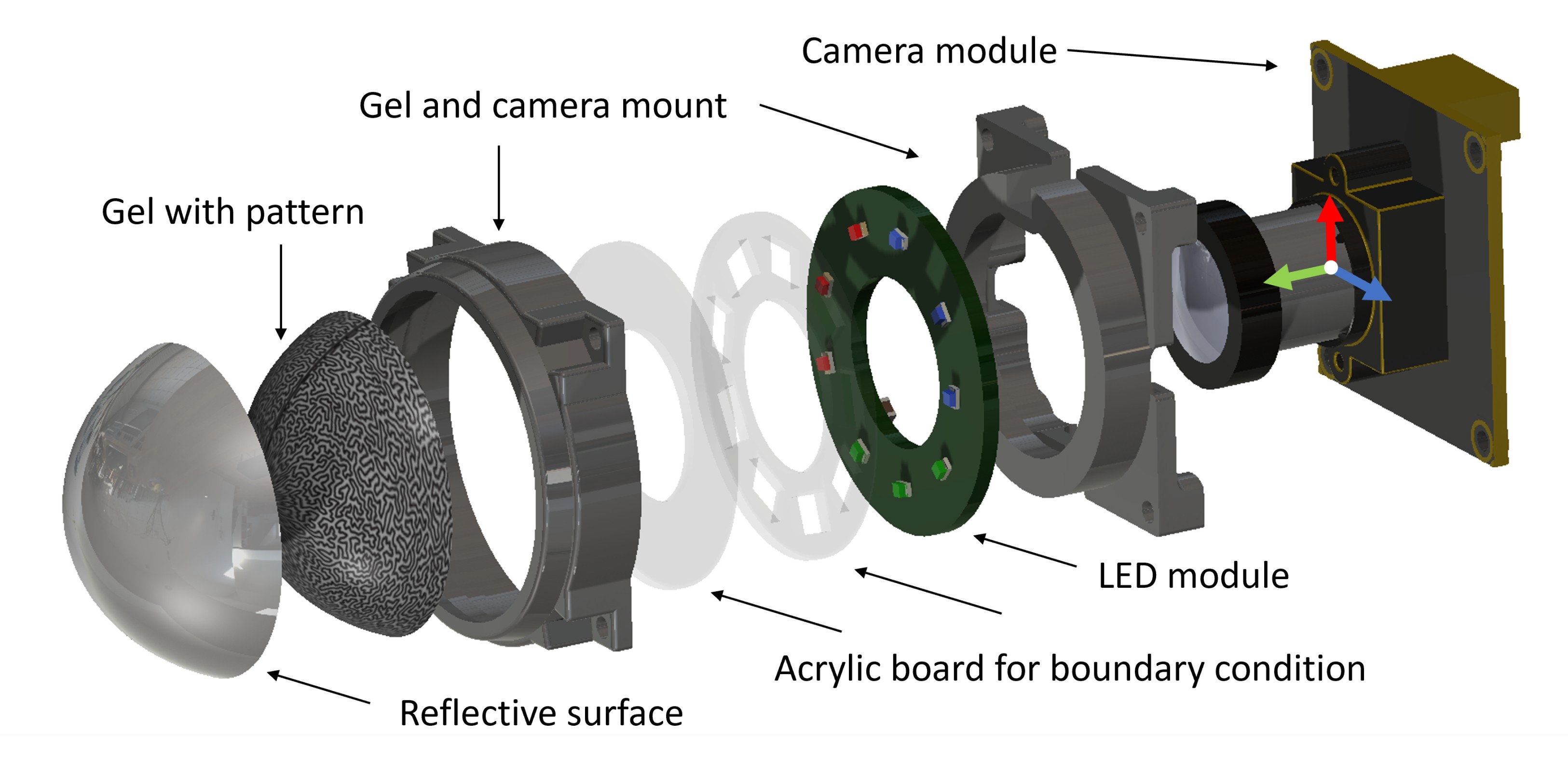}
% \vspace{-1em}
\caption{Proposed sensor design for TensorTouch. This modular layout shares the components common to general optical tactile sensors.}
\label{fig:newsensordesign}
\vspace{-1.4em}
\end{figure}

To verify the framework of TensorTouch, we verified each step of the framework - first, we designed an optical tactile sensor to verify the effectiveness of  TensorTouch, next, using the proposed framework, we collected and processed a dataset pair for training the network. Finally, we report various network architectures, including our proposed one, to estimate the stress tensor, displacement, and contact force from FEA. 

\subsection{Optical Tactile Sensor and Gel Configuration}

% \subsection{Improved sensor and gel design for reducing gap between simulation and actual sensors}

% \subsection{Generalization on multiple 3D-shaped optical tactile sensors}
% showing the structure of the sensor with different gel height, sensor figures, 
% gel construction, gel stiffness, gel spec what's improved, reason for making different configuration 
% whole structure of the sensor, LED, etc. 
% fabrication method is available 

% sensor spec 1600*1200, which can be always downsampled
% \subsection{Modular design of Optical Tactile Sensors for various type of the gel}

We propose a modular sensor design that enables easy simulation of different gel configurations without significant hardware modifications. This design reduces simulation complexity while improving accuracy by maintaining consistent boundary conditions across various sensor geometries.

As shown in Fig. \ref{fig:newsensordesign}, the sensor features a modular architecture where gel components can be easily swapped while keeping other parts unchanged. The design incorporates standard optical tactile sensor features with a 3D hemispherical shape, ensuring compatibility with existing sensor frameworks.

The gel requires full-surface attachment to the camera-LED module, which is crucial for maintaining consistent simulation boundary conditions common to most vision-based tactile sensors. While partial attachment is theoretically possible, full attachment ensures both sensor durability and simulation accuracy. Therefore, we fabricated a flat gel bottom surface that directly attaches to the lens and illumination components.

The illumination system consists of 9 LEDs arranged around a 220-degree fisheye lens, providing complete coverage of the sensor surface including edges. Clear acrylic boards prevent direct LED-gel contact while maximizing attachment area. The fisheye lens ensures comprehensive surface monitoring even during large deformations.

The two-part chassis design serves distinct functions: the inner chassis supports the sensor load and houses LED/acrylic components, while the upper chassis provides press-fit mounting for the camera-LED module. This separation enables modularity and allows easy gel replacement without affecting other components. Importantly, sensor loads transfer to the LED/acrylic components rather than directly to the camera, with the bottom chassis attached to the camera edge for structural support.
To reduce complexity, LEDs draw power from the camera's USB port, eliminating additional wiring requirements.

We tested different gel softness levels to demonstrate TensorTouch framework generalizability, with mechanical properties verified through testing (detailed in appendix). Gel attachment utilizes a combination of Loctite Power Grab adhesive and Sil-poxy for secure bonding. The flatter gel shape was selected over pure hemispherical geometry to demonstrate framework effectiveness across both flat and curved sensor configurations, showing broader applicability to various tactile sensor designs.

This modular approach enables researchers to adapt the framework to different sensor requirements while maintaining calibration accuracy and simulation fidelity.

% how we choose the softness of the gel -> softer gel has the pros and cons at the same time -> e.g. hard to apply the existing approach due to the large deformation, which is almost impossible to apply existing methodology such as ifem or others 
% Gel shape selection how - for the different types of gels - why and how? 

% \edit{more detailed description? e.g. how to attach the gel, adhesives, gel shape design, to measure the gel spec, etc. }

% \todo[inline]{Finish this}

\subsection{Simulation Result}

\begin{figure}[htbp]
    \centering
    \includegraphics[width=0.9\linewidth]{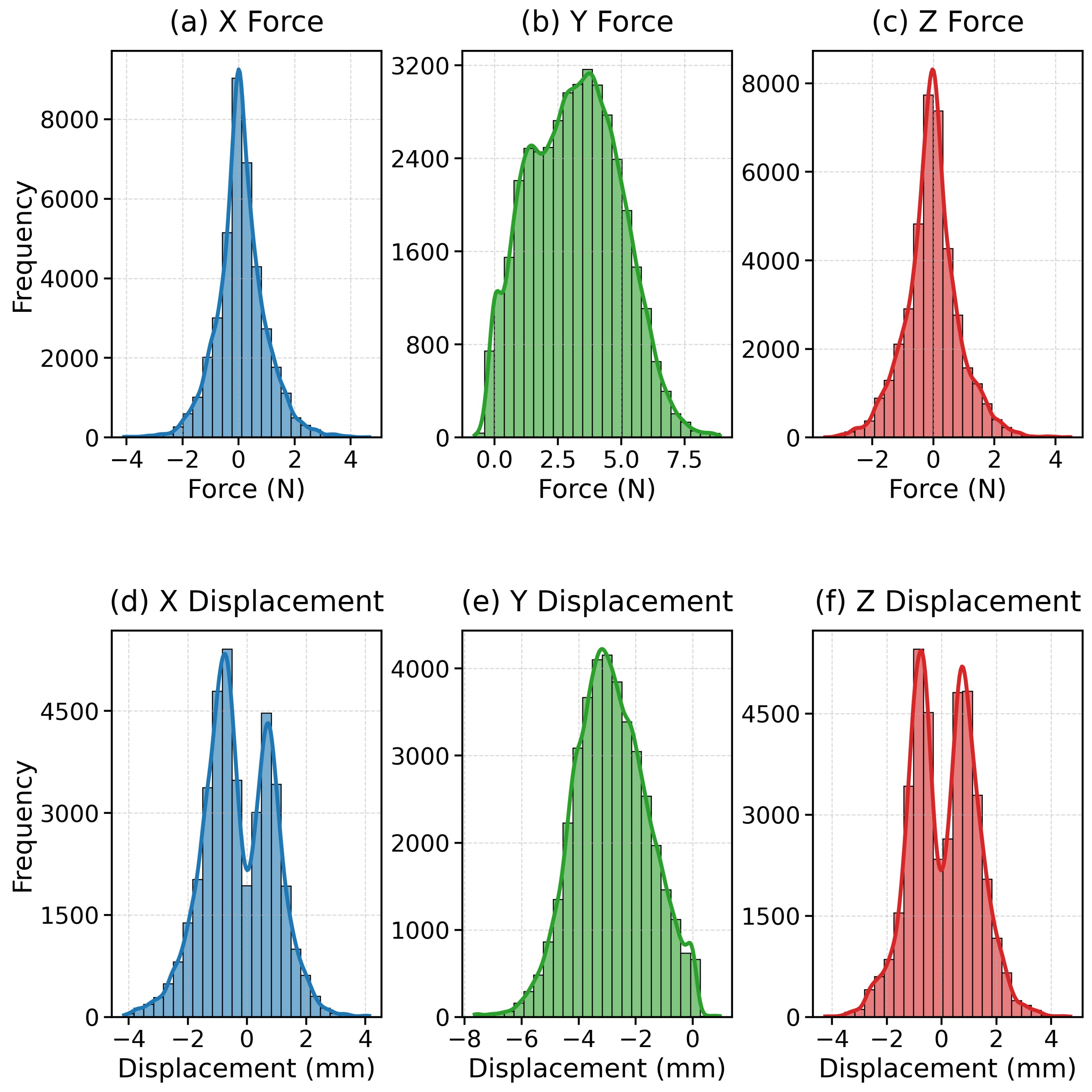}
    \caption{Distribution of force and displacement measurements of the dataset. The top row (a-c) shows force components in the X, Y, and Z directions measured in Newtons. 
    The bottom row (d-f) shows displacement components in the X, Y, and Z directions measured in millimeters.}
    \vspace{-1.2em}
    \label{fig:sensor_distributions}
\end{figure}

Fig.\ref{fig:sensor_distributions} shows the distribution of the collected and matched datasets. A total of 39,371 datapairs were collected for the sensor with extra-soft gel, while 38,079 datapairs were collected for the sensor with soft gel. Each dataset includes four sensors with the same gel softness. During data collection, approximately 7-9 trials were conducted, each lasting about five minutes. Every trial comprised around 100-200 indentation trajectories, each using a different indenter shape; some trials even involved more than one indenter.

After collecting the datasets and extracting keypoints for each indentation trajectory, finite-element simulations were run based on those keypoint poses. Simulating a single indentation trajectory required roughly 6-8 minutes - depending on the number of steps - to process and import all associated stress tensors, displacements, and contact-force data.

\subsection{Network Estimation Result}
% \subsection{pretraining effectiveness}

% \subsection{training result / estimation result}

% \todo[inline]{Finish this - result from network model will be set here, and also the result of estimation figure also required  }
\begin{table*}[htbp]
  \centering
  \setlength{\tabcolsep}{4pt}
  \small
  \begin{tabular}{@{}l c c c c c c@{}}
    \toprule
    \textbf{Network Structure} & \textbf{Params} &
    \textbf{PSNR} ($\uparrow$) &
    \textbf{SSIM} ($\uparrow$) &
    \textbf{MSE} ($\downarrow$) &
    \textbf{AUC$_{\mathrm{rel}}$} ($\uparrow$) &
    \textbf{FPR$_{\mathrm{rel}}$} ($\downarrow$) \\
    \midrule
    Hiera with Q‐upsampling & \textbf{76\,M} & \textbf{55.9756} &\textbf{ 0.99066} & \textbf{2.76\,$\times10^{-6}$} & \textbf{0.85499} & \textbf{0.01852 }\\
    DenseNet               & 80\,M & 53.9215 & 0.98936 & 4.12\,$\times10^{-6}$ & 0.82255 & 0.02501 \\
    ResNet                 &113\,M & 52.0419 & 0.98783 & 6.26\,$\times10^{-6}$ & 0.80451 & 0.02896 \\
    ViT Encoder            & 114\,M & 46.8280 & 0.98831 & 2.09\,$\times10^{-5}$ & 0.82149 & 0.02650 \\
    ViT DPT                & 144\,M & 44.2347 & 0.82281 & 3.79\,$\times10^{-5}$ & 0.08199 & 0.21833 \\
    Hiera with Vit Decoder & 191\,M &46.9025 & 0.98942 & 2.05\,$\times10^{-5}$ & 0.785961 & 0.037147 \\
    \bottomrule
  \end{tabular}
  \caption{Performance comparison of different network architectures.}
  \label{table:network_comparison}
\end{table*}

Using the collected dataset, an ablation study is conducted across various network architectures to verify the effectiveness of stress tensor estimation from sensor images. As shown in Table \ref{table:network_comparison}, the proposed Hiera-based model is compared with several baselines including DenseNet, ResNet, and different Vision Transformer configurations. 

While the proposed model consists of Hiera with Q-upsampling architecture to perform cross-attention between different resolution levels to generate dense pixel-wise predictions, the other baselines show existing effective network models. The DenseNet baseline, based on the DenseNet-161 architecture \cite{huang2017densely}, has demonstrated effectiveness in sensor deformation estimation \cite{doDT2.0} and uses densely connected convolutional blocks with skip connections to preserve fine spatial information. The ResNet baseline uses the ResNet-152 backbone, \cite{he2016deep} which contains residual connections. We also evaluated a standard Vision Transformer (ViT) encoder \cite{dosovitskiy2020vit} without the sophisticated upsampling mechanisms. The last evaluation baseline consists of ViT encoder with Dense Prediction Transformer (DPT) decoder \cite{ranftl2021vision} combination to show the effectiveness of the selected decoder.

All baseline models were trained under the same conditions to ensure fair comparison. We employed the AdamW optimizer with a learning rate of 0.001 with a linear cosine annealing schedule with 500 warmup steps. Models were trained with L1 loss for 200-250 epochs depending on convergence, while choosing best model to avoid overfitting. The batch sizes are set to 64 for most models and 96 for ResNet due to memory requirements. For the Hiera/ViT models, we perform the same pretraining process with masked autoencoder reconstruction \cite{MAE}, which has proven to lead to effective representations for downstream tasks, similar to T3 and Hiera\cite{zhao2024transferable, Hiera}, before fine-tuning on the stress tensor estimation. During pretraining, 75\% of the input patches from 6-channel are randomly masked.

The reported metrics in Table \ref{table:network_comparison} represent averaged performance across all five output modalities produced by each network head - displacement, normal and shear stress distributions, and contact normal / shear force distributions. Each model generates these five distinct images simultaneously and the metrics computed individually for each modality and then averaged to each metrics. The proposed metrics include Peak Signal-to-Noise Ratio (PSNR), structural similarity index measure (SSIM), and mean-squared error (MSE). The FPR\_rel metric measures the fraction of samples that relative error is more than 100\% of the ground-truth value, i.e. the fraction of predictions that are off by more than the true value. The AUC\_rel metric measures the area under the curve for relative error behavior, indicating how well the model performs across different accuracy requirements. A higher AUC\_rel suggests the model maintains good performance across the full range of relative error tolerances.  

DenseNet, despite having comparable parameter count (80M), achieves significantly lower performance with PSNR of 53.92 and AUC\_rel of 0.82255. The ResNet architecture, with the highest parameter count (113M), shows even more limited performance (PSNR: 52.04, AUC\_rel: 0.80451), suggesting that simply increasing network depth does not effectively address the challenges of tactile force field estimation.

However, the ViT DPT baseline performs poorly across all metrics, particularly in SSIM (0.82281) and AUC\_rel (0.08199), with a very high FPR\_rel (0.21833) because of false positive estimation on shear stress measurement. Furthermore, Hiera with vanilla ViT decoder underperforms our Q-Upsampling approach across all metrics despite using 191M parameters versus 76M. This indicates that while transformer architectures show promise, \textbf{the specific design of the decoder and multi-scale feature fusion is critical for dense, multi-channel prediction tasks in the tactile domain}. Hiera with Q-Upsampling approach demonstrates the best performance from the overall metrics. 

% \begin{table*}[htbp]
%   \centering
%   \setlength{\tabcolsep}{4pt}
%   \small
%   \begin{tabular}{@{}l c c c c c c@{}}
%     \toprule
%     \textbf{Network Structure} & \textbf{Params} &
%     \textbf{PSNR} ($\uparrow$) &
%     \textbf{SSIM} ($\uparrow$) &
%     \textbf{MSE} ($\downarrow$) &
%     \textbf{AUC$_{\mathrm{rel}}$} ($\uparrow$) &
%     \textbf{FPR$_{\mathrm{rel}}$} ($\downarrow$) \\
%     \midrule
%     Hiera with Q‐upsampling & \textbf{76\,M} & \textbf{55.9756} &\textbf{ 0.99066} & \textbf{2.76\,$\times10^{-6}$} & \textbf{0.85499} & \textbf{0.01852 }\\
%     DenseNet               & 80\,M & 53.9215 & 0.98936 & 4.12\,$\times10^{-6}$ & 0.82255 & 0.02501 \\
%     ResNet                 &113\,M & 52.0419 & 0.98783 & 6.26\,$\times10^{-6}$ & 0.80451 & 0.02896 \\
%     ViT Encoder            & 114\,M & 46.8280 & 0.98831 & 2.09\,$\times10^{-5}$ & 0.82149 & 0.02650 \\
%     ViT DPT                & 144\,M & 44.2347 & 0.82281 & 3.79\,$\times10^{-5}$ & 0.08199 & 0.21833 \\
%     \bottomrule
%   \end{tabular}
%   \caption{Performance comparison of different network architectures.}
%   \label{table:network_comparison}
% \end{table*}

\begin{figure*}[t]
    \centering
    \includegraphics[width=0.92\textwidth]{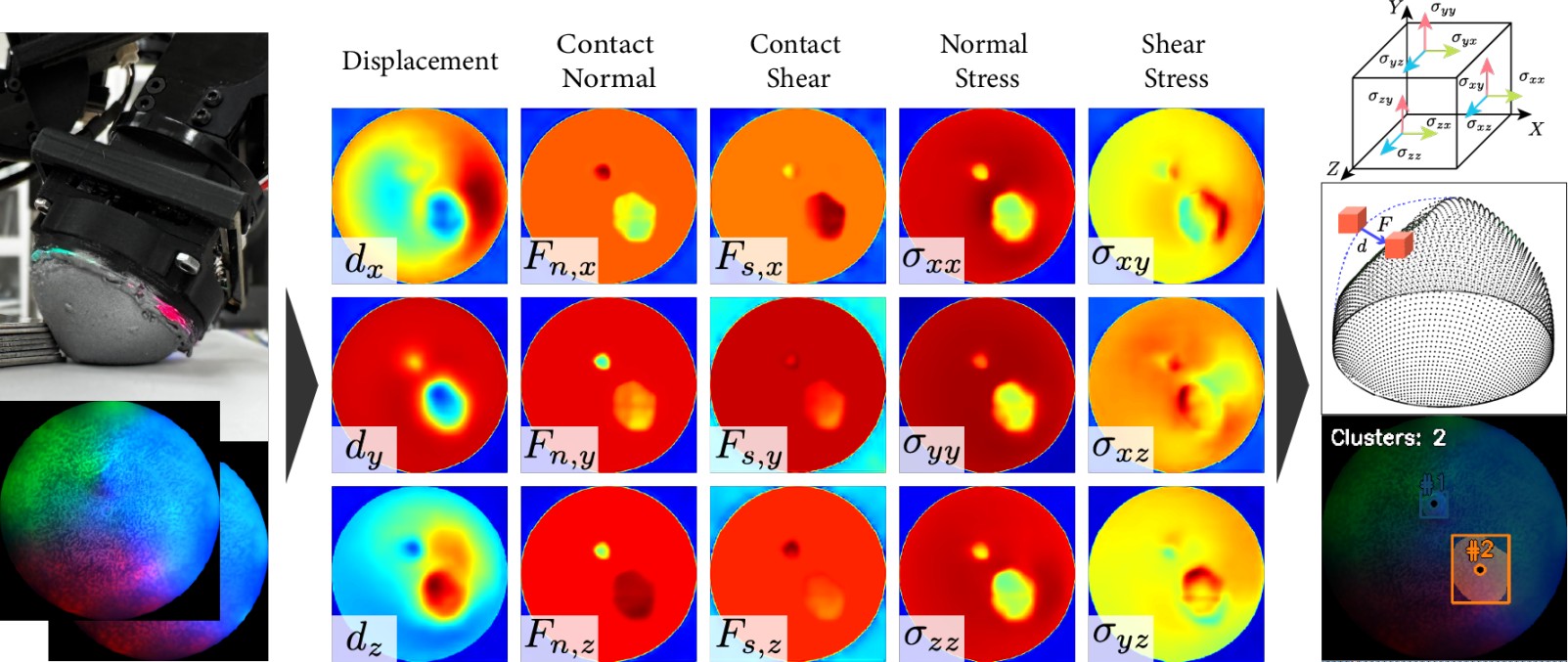}
    \caption{Visualization of multi-channel tactile sensing outputs generated by our calibrated model from a single tactile image (top right). The left panels show predicted physical quantities across three coordinate axes (X, Y, Z from top to bottom): displacement fields, normal stress distributions, contact normal forces, contact shear forces, stress tensor components, and integrated contact forces. The bottom right image demonstrates successful identification of two distinct contact regions with their corresponding force magnitudes and centers, enabling multi-contact tracking.}
    \vspace{-4mm}
    \label{fig:tactile_outputs}
\end{figure*}

\subsection{Qualitative Result of TensorTouch}

Fig. \ref{fig:tactile_outputs} demonstrates the comprehensive multi-channel tactile sensing capabilities of our calibrated model through a representative interaction between a planar surface and a card-like object, resulting in two distinct contact patches. Each column in the visualization grid represents different physical quantities extracted from a single tactile image: displacement fields ($d_x, d_y, d_z$), contact normal forces ($F_{n,x}, F_{n,y}, F_{n,z}$), contact shear forces ($F_{s,x}, F_{s,y}, F_{s,z}$), normal stress components ($\sigma_{xx}, \sigma_{yy}, \sigma_{zz}$), and shear stress components ($\sigma_{xy}, \sigma_{xz}, \sigma_{yz}$). The relationship between contact forces and stress tensors reflects fundamental continuum mechanics principles: contact forces represent the integrated effect of stress distributions over the contact area, while stress tensors capture the internal material response at each point. Specifically, contact normal and shear forces are derived from the corresponding stress components multiplied by the local contact area and appropriate offset ratios, explaining why both quantities exhibit similar spatial patterns but with contact forces showing more localized, concentrated distributions at the actual interface regions.

The spatial distribution of these quantities reveals important tactile sensing characteristics, particularly how contact forces naturally encode contact area information. In regions without physical contact, both force and stress values approach zero, appearing as orange or red coloring in the relative intensity visualization. This inherent contact area encoding makes the force estimates particularly valuable for manipulation tasks, as they directly indicate where and how strongly objects are interacting with the sensor surface. The right panels of the figure demonstrate the 3D interpretation capability through our 2D-3D correspondence mapping, showing how the estimated quantities can be projected into a point cloud representation where each point contains the full stress tensor and displacement information encoded as RGB values. This visualization confirms that our framework successfully captures the complex multi-contact scenario with two distinct interaction regions, each characterized by its own force magnitude, direction, and contact geometry, enabling the precise multi-contact tracking essential for advanced dexterous manipulation tasks.
% \subsection{Evaluation result}

% \begin{figure}[htb]
% % \vspace{-1em}
% \centering
% \includegraphics[width=0.9\linewidth]{Figures/meshexample.jpg}
% % \vspace{-1em}
% \caption{New sensor design should be here}
% \label{fig:meshex}
% % \vspace{-2em}
% % 
% \end{figure}

\section{Evaluation}

The evaluation of sensor has been conducted in two directions - first, the position and force of cluster from the sensor are compared with the actual position and force collected from motion capture system and ATI F/T sensor. Next, to verify the usefulness of the sensor in the dexterous manipulation, we build a corner case that handles multiple deformable objects, specifically pinching two strings with a pair of optical sensors attached on a multi-finger robot hand. 

\subsection{Evaluation and Measurement Comparison with Clusterization and Mocap System}

To demonstrate the effectiveness of the sensor, its position and force outputs were compared with actual values captured from the data collection system. To evaluate force information, the ATI sensor measurements were compared with the sensor outputs. The ATI sensor measurements were transformed into the global frame using the pose of the indenter. During the evaluation, the focus was on single-contact scenarios, although this approach can be extended to multiple indenters.

To evaluate contact point accuracy, the contact cluster position derived from the sensor was utilized in relation to the indenter pose. First, the center position of each cluster in 2D image coordinates was extracted and converted into 3D positions in the real-world coordinate system. These cluster positions were then compared with the indenter poses. It is important to note that the contact cluster position does not exactly match the pose of the indenter, since the origin of the indenter is fixed at its bottom center, while the actual contact position varies based on the indenter's geometry. Even when using hemispherical indenters to minimize this discrepancy, some position differences are inherently expected.

\begin{figure}[t]
    \centering
    \subfloat[][\label{fig:force_result}]{\includegraphics[width=0.47\linewidth]{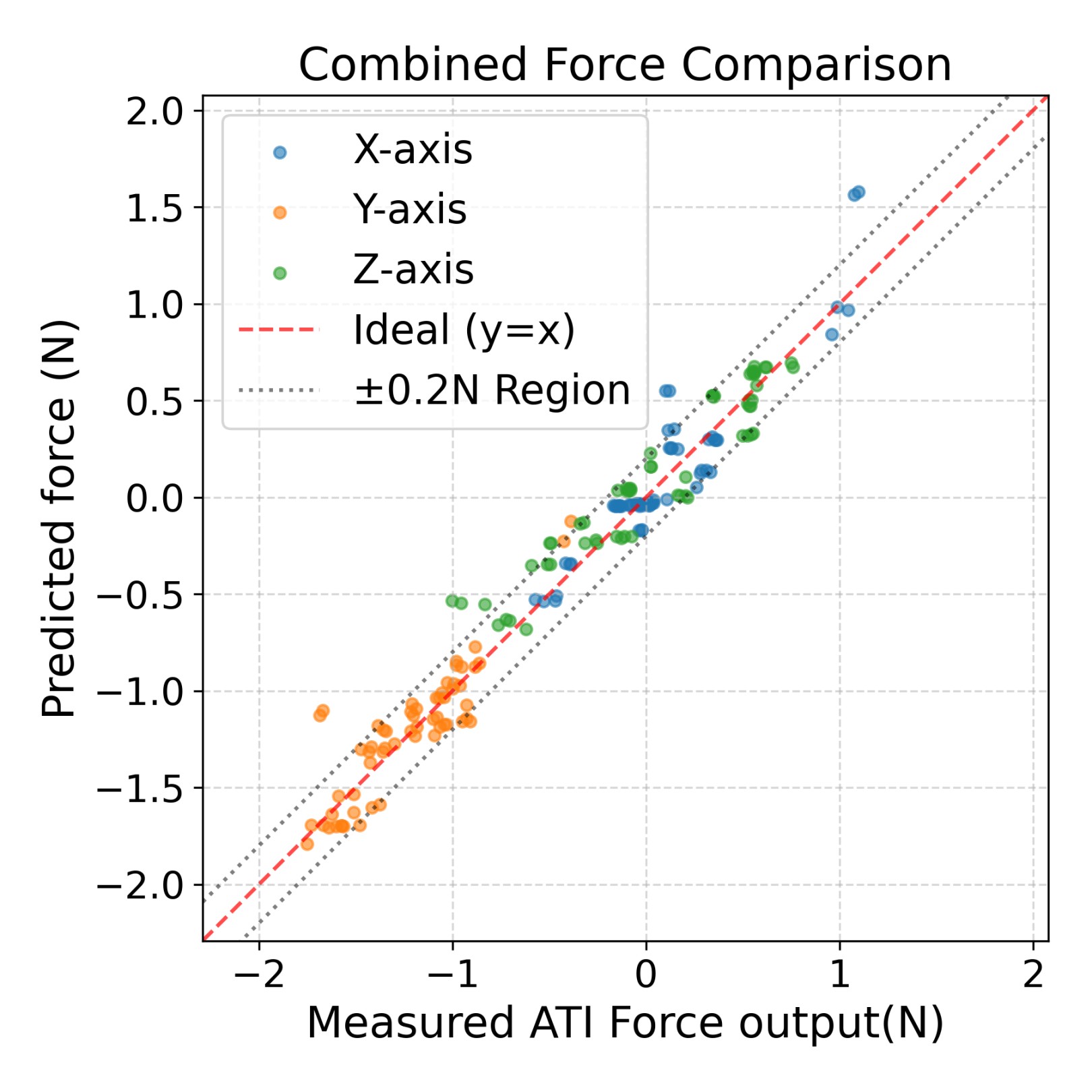}}
    \hfill
    \subfloat[][\label{fig:pos_result}]{\includegraphics[width=0.47\linewidth]{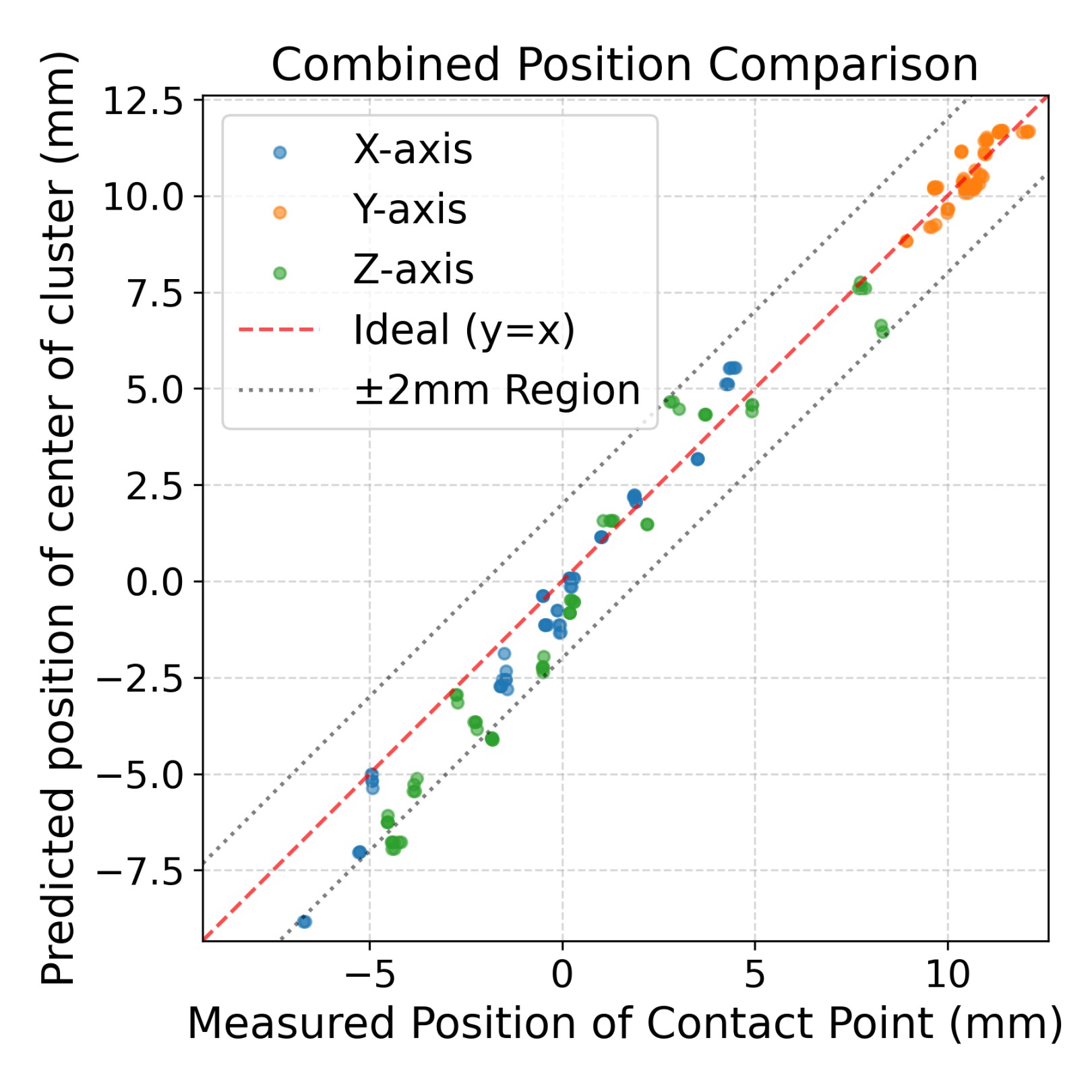}}
    \caption{Evaluation result of the sensor. (a) Measured ATI force sensor output vs predicted force; (b) measured position of tip of indenter via motion capture system vs predicted position of the center of clusterized indentation.}
    \vspace{-1.7em}
    \label{fig:forcepos_result}
\end{figure}

60 data points were collected for sensor evaluation. The results of force and position comparisons are shown in Fig. \ref{fig:forcepos_result}. The x-axes in both figures represent the measured position and force information recorded from the motion capture system and ATI sensor, while the y-axes show the predicted values of the position of the center of cluster and forces. Both results demonstrate good alignment in all three spatial directions (x, y, and z). The mean position errors in the x, y, and z coordinates are 0.684, 0.376, and 1.292 mm, respectively. The mean force errors are 0.106, 0.113, and 0.139 N in the x, y, and z directions, with an overall force magnitude error of 0.239 N. These results confirm that the contact force clusterization and position estimation methods closely match the actual measurements from the motion capture and ATI sensor systems. Please note that the inherent resolution of the motion capture system is less than 0.2 mm, and the resolution of the ATI-mini SI-145-5 force/torque sensor is 0.0625 N.

% Position Error Mean: 0.6837092531764548, 0.37642640624259743, 1.292155407765097 mm
% Force Error Mean: 0.10596352354664139, 0.11292847680930188, 0.1389024506476296 N
% Force Error Mean: 0.23926650573557995 N
\subsection{Evaluation with Multi-object Manipulation}
% With two strings, 

Advanced dexterous manipulation requires the ability to distinguish between multiple simultaneously contacted objects and respond selectively to changes in their states. To demonstrate the practical utility of TensorTouch's calibrated stress tensor estimation, a challenging multi-object manipulation scenario was designed that tests both the force discrimination capabilities and the precision control enabled by the rich tactile feedback.

The experimental task involves simultaneously contacting two deformable objects (strings or cables) with tactile sensor-equipped fingertips, detecting differential motion between the objects, and maintaining selective contact with only the moving object. This scenario mirrors real-world manipulation challenges where robots must identify and respond to specific objects among multiple contacts, such as untangling cables or selectively manipulating individual strands in textile handling.

The task evaluation employed various cable and string combinations with different material properties including varying stiffness, thickness, and surface texture. Performance was assessed based on the system's ability to accurately classify which object experienced greater applied force and subsequently maintain grasp on only the target object through coordinated finger repositioning. This evaluation framework tests both the sensing accuracy of the force field estimation and the effectiveness of the tactile-feedback-driven control system.

% I tried rephrasing, but I'm I may be getting this confused with the next section 

% Just as humans can detect subtle variations in string tension when tying shoes, our method aims to replicate this touch-based dexterity. In our experimental setup, two strings are positioned between a pair of tactile sensors.

% We tested the system using various cables and strings with different hardness levels and thicknesses. Success was defined by two criteria: first, whether the sensor could accurately identify which string experienced greater applied force, and second, whether the sensor could subsequently grasp only the targeted string based on its output signals.

\subsubsection{Graph-Based Contact Control for Multi-Object Manipulation}

To enable precise finger positioning based on tactile feedback, a graph-based control framework was implemented that leverages the calibrated force and displacement estimates from TensorTouch. This approach addresses the challenge of controlling finger contacts in scenarios where analytical models of multi-object interaction dynamics are complex or unavailable.

The contact graph construction process involves systematic exploration of the joint configuration space while maintaining contact between the thumb and index fingers. Data collection captures pairs of joint configurations alongside corresponding tactile responses, specifically recording joint angles for both fingers and contact cluster positions detected by the thumb and index-mounted tactile sensors. The resulting dataset encompasses approximately 2,700 contact configurations spanning the operational workspace.

Graph connectivity is established by linking configurations that are physically reachable through small joint displacements, with edge weights determined by joint space distances. This connectivity structure enables identification of smooth transition paths between different contact states while respecting the kinematic constraints of the hand. The dense sampling across the workspace ensures robust path planning capabilities even in complex contact scenarios.

During manipulation tasks, the system locates the graph node closest to the current joint configuration and contact state, then computes optimal paths to desired contact configurations using Dijkstra's shortest path algorithm. This approach enables stable navigation through contact-rich regions of the state space without requiring precise analytical models of the finger-object interaction dynamics, which can be particularly challenging when dealing with deformable objects like strings and cables.

% \edit{We included the distribution of contact graph visualization with }

\subsubsection{Multi-Object String Manipulation Evaluation}

The evaluation protocol tests the integrated performance of force field estimation and graph-based control through a systematic multi-stage manipulation task. The experimental setup positions two objects between the thumb and index fingers, each equipped with calibrated tactile sensors as shown in Fig. \ref{fig:main}, where TensorTouch processes tactile deformation images to generate comprehensive contact information during the manipulation sequence.

The manipulation protocol begins with simultaneous contact establishment between both fingertips and both target objects, creating a stable initial configuration with multiple contact regions. Force monitoring through the contact clustering algorithm tracks the magnitude and spatial distribution of forces across both sensor surfaces. When differential motion is introduced by gently displacing one object, the clustering system detects asymmetric force changes indicating which object is experiencing external perturbation.

Upon detection of differential motion, the control system queries the contact graph to identify valid trajectories for repositioning the fingers to maintain contact solely with the moving object. Path planning considers both the desired contact cluster positions and the kinematic constraints encoded in the graph structure, ensuring stable finger repositioning while maintaining appropriate contact forces throughout the transition.

Object combinations tested include pairs of identical cables to assess discrimination sensitivity, different cable types to evaluate performance with varying material properties, cable-rigid object combinations to test mixed compliance scenarios, and pairs of rigid objects as a control condition. Each combination presents unique challenges for the force discrimination and control systems, with performance metrics capturing both detection accuracy and manipulation success rates.

\begin{table}[htbp]
\setlength{\tabcolsep}{10pt}
\centering
\begin{tabular}{lccc}
\toprule
\textbf{Object} & \textbf{Success} & \textbf{False} & \textbf{Other} \\
\textbf{Combination} & \textbf{Rate} & \textbf{Detection} & \textbf{Failures} \\
\midrule
Two identical & 21/31 & 4/31 & 6/31 \\
cables (A+A) & (67.7\%) & (12.9\%) & (19.4\%) \\
\midrule
Different cables & 15/20 & 2/20 & 3/20 \\
(A+B) & (75.0\%) & (10.0\%) & (15.0\%) \\
\midrule
Cable + & 17/20 & 1/20 & 2/20 \\
rigid object & (85.0\%) & (5.0\%) & (10.0\%) \\
\midrule
Two rigid & 18/20 & 0/20 & 2/20 \\
objects & (90.0\%) & (0.0\%) & (10.0\%) \\
\bottomrule
\end{tabular}
\caption{Success rates for selective grasping across object combinations. Cable A refers to 2.18mm thickness cables, and B refers to the 3.01mm thickness cables. For the rigid objects, we tried with the pens with 8mm diameter.}
\vspace{-1em}
\label{table:evaluation}
\end{table}

The experimental results demonstrate reliable distinction between multiple contact regions and successful selective grasp maintenance across diverse object combinations. Performance shows clear correlation with the mechanical differences between objects, achieving highest success rates when manipulating objects with distinct stiffness properties. The system successfully handles the challenging case of identical cables, indicating sufficient force discrimination resolution for subtle material variations.

Failure analysis reveals that most unsuccessful trials result from insufficient joint torque in the Allegro Hand rather than sensing or control algorithm limitations, suggesting that the tactile sensing and path planning components perform robustly within the hardware constraints. The evaluation confirms that calibrated force field estimates provide sufficient spatial and temporal resolution for complex manipulation tasks involving multiple deformable objects, validating the practical utility of the TensorTouch framework for advanced dexterous manipulation applications.

% table with success rate and failure mode (wrong movement, wrong classification, not moving, and consistency - when does it moved)

% trial with two identical cables - probe cables (for 20 trial, 14 success, 6 failure - 3 bc of not moving (consistency), 1 classification error, 2 wrong movement)
% trial for different cable 

% \subsection{Evaluation setup - card holder }
% franka arm + allegro hand, also share leap-hand setup 
% analytical approach of using this setup d

% \subsection{estimation of force closure of known polytope}

% Grasping object with multi finger, while keep recording the force at the fingertip 

\section{Conclusion}

This paper presented TensorTouch, a comprehensive framework for stress tensor estimation from 3D optical tactile sensors designed specifically for complex manipulation tasks. We developed a physics-based calibration methodology that combines motion capture data collection, FE analysis, and deep learning to extract detailed contact information from tactile images. Our approach enables accurate estimation of contact area, normal and shear forces, deformed shape, and stress distributions across the sensor surface, even under large deformations. Experimental validation demonstrated that our framework achieves sub-millimeter position accuracy and precise force estimation across multiple sensor configurations, enabling reliable multi-contact tracking during complex manipulation tasks.

The results demonstrate significant advantages for dexterous manipulation in contact-rich environments. By extracting comprehensive stress tensors from vision-based tactile sensors, robots can better understand and respond to complex force interactions during manipulation. Our graph-based control framework leveraging these rich tactile features enabled selective string grasping with success rates of up to 90\% in challenging multi-object scenarios. The modular sensor design and generalizable calibration approach presented in this work provide a foundation for integrating sophisticated tactile sensing into various robotic platforms without requiring custom sensor designs for each application. This bridges an important gap between the mechanical properties of highly compliant, sensitive tactile sensors and their practical utility in real-world robotic systems.

Future work will focus on extending this framework in three key directions. First, we aim to generate realistic images of sensor deformation directly from indenter pose and shape information, potentially enabling physics-based data augmentation for training. Second, we plan to integrate our entire modeling pipeline into simulated environments to accelerate policy training for complex manipulation tasks, addressing the well-known sample efficiency limitations of reinforcement learning. Finally, we will investigate the effectiveness of TensorTouch in behavior cloning frameworks that leverage the rich stress tensor information rather than relying solely on raw RGB images from tactile sensors. This approach could enable more data-efficient learning of skilled manipulation behaviors by providing structured, physically meaningful representations of contact interactions.

\trorev{\section{Acknowledgments}
\noindent Authors thank to Skyler R. St. Pierre and Prof. Ellen Kuhl for borrowing biaxial and uniaxial testing equipment for measuring mechanical property of the gel.
}

\section{Appendix}

\subsection{Analysis on Accuracy of Calibration System}

To validate our motion capture system's accuracy for physical object tracking, we conducted error analysis using a CNC machine as ground truth reference. The experimental setup consisted of a CNC machine with a rigid body marker array attached to its movable head and a stationary reference marker array. The CNC machine was programmed to move in precise 1 mm increments along each axis (X, Y, Z) across its operational range (190 mm in X-axis, 36 mm in Y-axis, and 110 mm in Z-axis).

At each position, after allowing the system to stabilize, we recorded both the commanded CNC position and the relative transform between marker arrays as reported by the motion capture system. The absolute difference between the mocap-measured displacement and the known CNC displacement was calculated to quantify tracking error. This process generated comprehensive error data across the full workspace, enabling statistical analysis of system accuracy.
Our results demonstrate sub-millimeter accuracy with mean errors of 0.0849 mm (SD = 0.0524 mm), 0.0405 mm (SD = 0.0230 mm), and 0.0308 mm (SD = 0.0291 mm) for X, Y, and Z axes respectively. The position-dependent error analysis shows no strong correlation between error magnitude and position, indicating consistent tracking accuracy throughout the workspace. These error values confirm that our motion capture system provides sufficient accuracy for establishing reliable correspondence between simulation and real-world object poses.

\begin{figure}[h]
\centering
\includegraphics[width=0.45\textwidth]{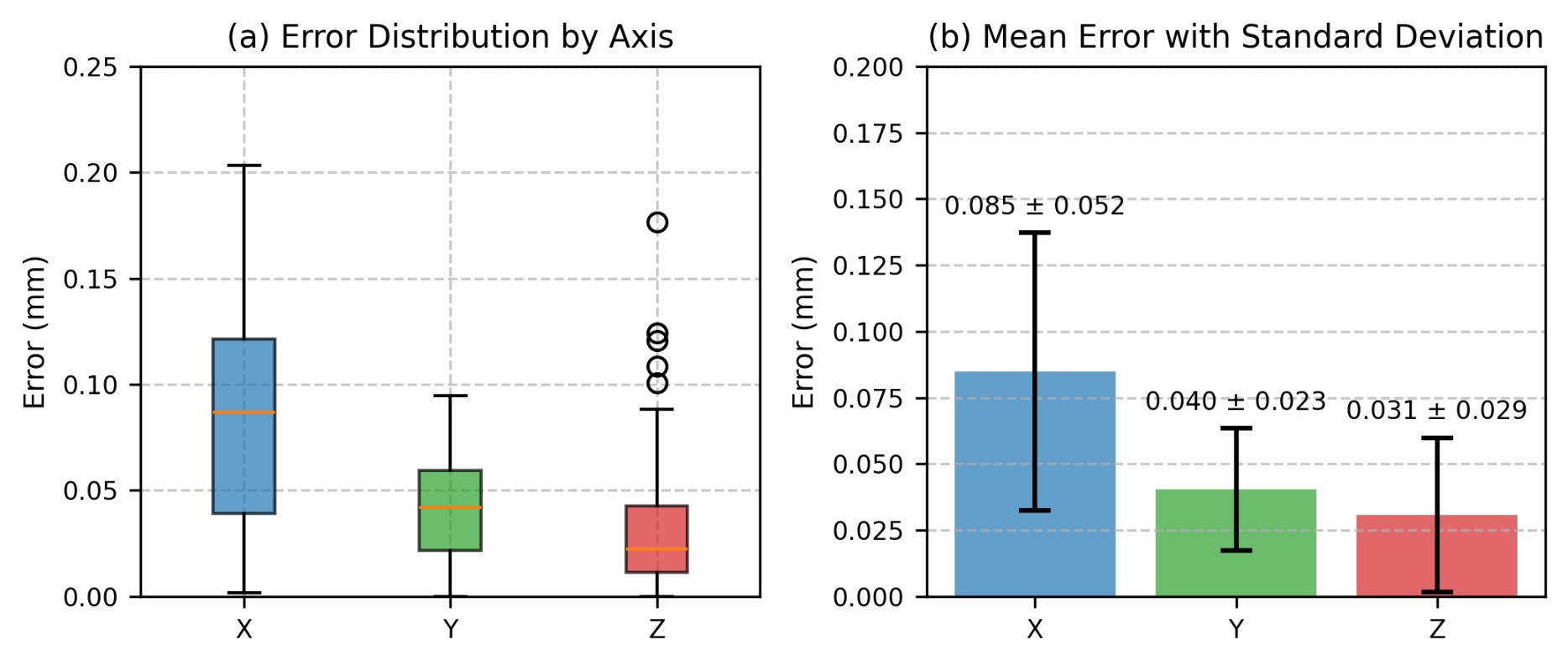}
\caption{Error distribution of the motion capture system across the three spatial axes. (a) Boxplot representation showing the median, quartiles, and outliers of positional errors for each axis. (b) Mean errors with standard deviation error bars, highlighting the sub-0.1 mm accuracy achieved in all directions.}
\vspace{-1.7em}
\label{fig:mocap_error_stats}
\end{figure}

\begin{figure}[h]
\centering
\includegraphics[width=0.45\textwidth]{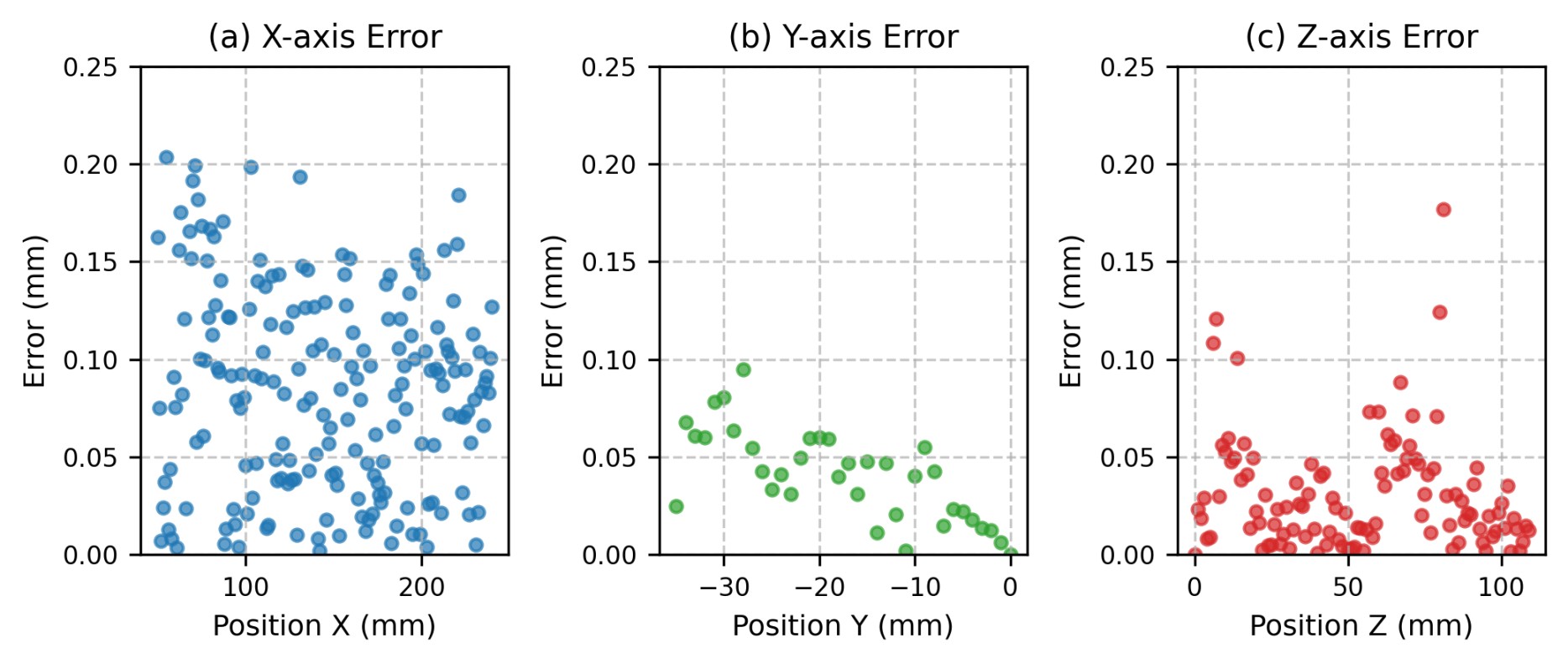}
\caption{Position-dependent error analysis for the motion capture system. Scatter plots showing the relationship between measured error and position for (a) X-axis, (b) Y-axis, and (c) Z-axis. The relatively uniform distribution of errors across positions indicates that tracking accuracy remains consistent throughout the workspace, with no significant systematic biases or position-dependent error patterns observed in any dimension.}
\vspace{-1.5em}
\label{fig:mocap_err_vs_pos}
\end{figure}

\subsection{Characterization of Gel Materials Mechanical Properties}

The mechanical properties of the silicone elastomers used in tactile sensors were characterized using a CellScale BioTester 5000, a specialized equipment for soft material testing. This characterization was critical for creating accurate finite element models that could properly simulate large deformations of the gel surfaces during contact.

\subsubsection{Sample Preparation and Testing Methodology}

Three silicone elastomer formulations were prepared with varying activator-to-base ratios: 15:1 (hard), 20:1 (soft), and 23:1 (extra soft). For each formulation, five identical 8mm × 8mm × 1mm specimens were fabricated and mounted on the CellScale BioTester using rakes with 0.7mm tine spacing for uniform force distribution. The testing apparatus included a temperature-controlled chamber at 25°C and high-resolution cameras for real-time strain measurement using digital image correlation.
Each specimen underwent uniaxial and biaxial testing protocols. Uniaxial tests stretched samples in the x-direction with free y-direction contraction, then were repeated in the y-direction. Samples were subjected to three complete load-unload cycles to capture potential Mullins effect (stress softening), reaching maximum strains of approximately 50\%—representative of deformation ranges in typical tactile sensor manipulation tasks.

Biaxial tests simultaneously stretched samples in both x and y directions at identical rates to achieve equibiaxial strain conditions. This testing mode is particularly relevant for tactile sensors as it better represents complex loading conditions during multi-directional contacts. Similar to uniaxial tests, three complete loading cycles ensured mechanical response stability.

\subsubsection{Data Processing and Material Model Fitting}

Force-displacement data from all tests was converted to engineering stress-strain curves and averaged across all five specimens per formulation. For uniaxial tests, results from both directions were combined to create comprehensive datasets accounting for material anisotropy. The data exhibited classic hyperelastic behavior with nonlinear stress-strain relationships and minimal hysteresis, confirming the appropriateness of hyperelastic constitutive models in FE analysis.

Experimental data was fitted to the Yeoh hyperelastic model, particularly suitable for elastomers undergoing large deformations. The processed stress-strain data and corresponding Yeoh model parameters were incorporated into FE analysis, enabling accurate modeling of gel mechanical response during simulated indentation tests and capturing complex deformation patterns critical for vision-based tactile sensor calibration. Experimental characterization confirmed that increasing crosslinker-to-base ratio systematically produces softer materials, with the 23:1 formulation exhibiting approximately one-fourth the stiffness of the 15:1 formulation.

\begin{figure}[t]
    \centering
    \subfloat[][\label{fig:silicone_measure}]{\includegraphics[width=0.40\linewidth]{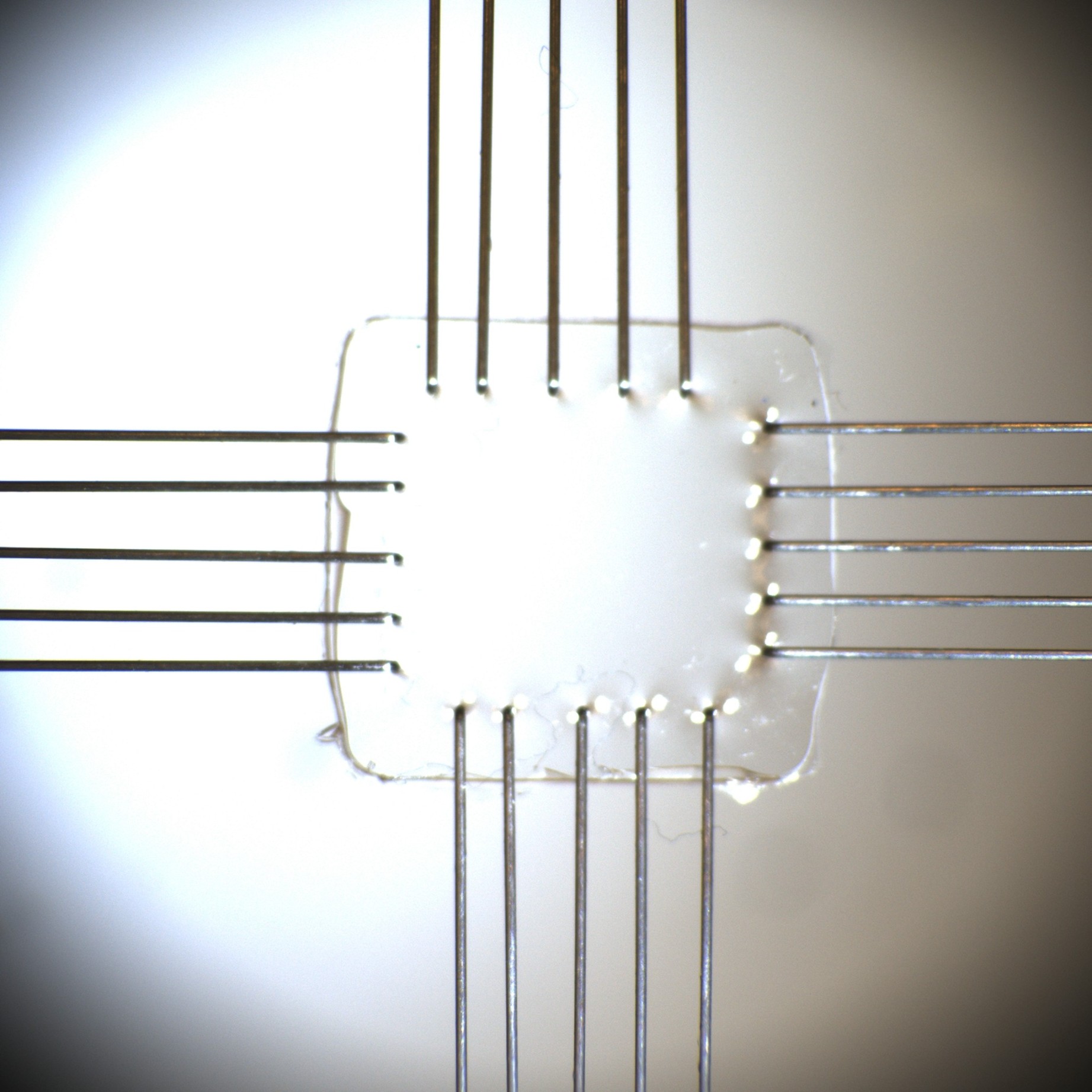}}
    \hfill
    \subfloat[][\label{fig:gel_comparison}]{\includegraphics[width=0.56\linewidth]{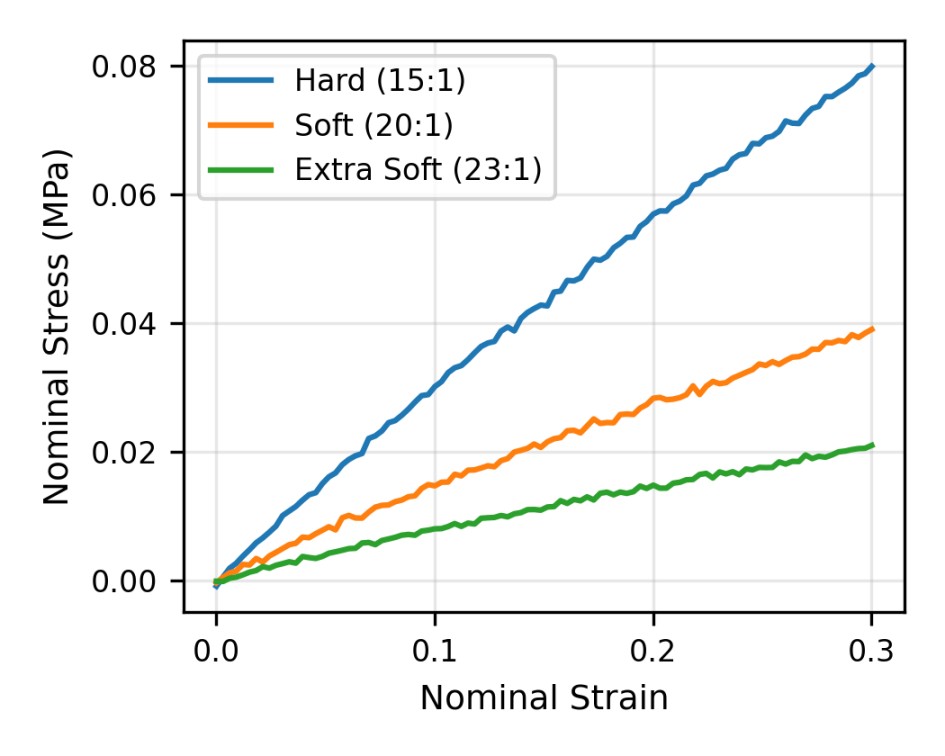}}
    \caption{Mechanical characterization of silicone elastomers for tactile sensors. (a) a setup with a square silicone specimen mounted using rake attachments for biaxial testing; (b) Uniaxial stress-strain curves showing the effect of crosslinker-to-base ratio on mechanical properties.}
    \label{fig:mechanical_characterization}
\end{figure}

\subsection{Implementation Details of Optimal Fingertip Pose Selection}
\begin{figure}[t]
    \centering
    \includegraphics[width=0.45\textwidth]{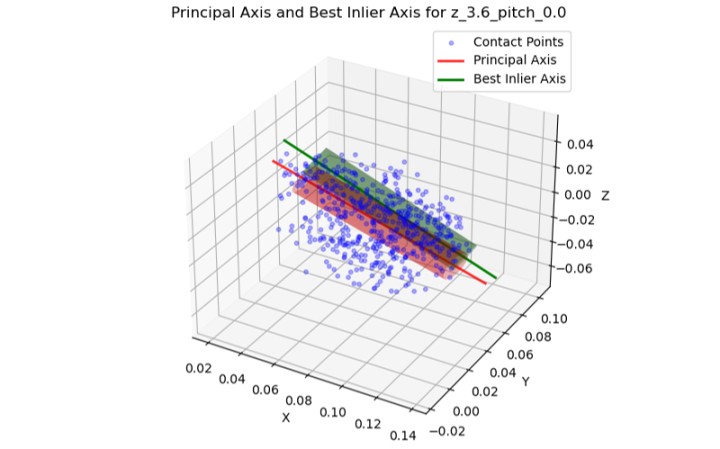}
    \caption{Contact points (blue) in 3D space for a representative configuration ($z=4.3\,\mathrm{cm}$, pitch $=0.0^\circ$). The principal axis is shown in red, and the best inlier axis (from the RANSAC-like cylinder fitting) is shown in green.}
    \label{fig:principal_axis_and_best_inlier_axis}
    \vspace{-1.7em}
\end{figure}

We developed a systematic pipeline for selecting the thumb's distal-link pose to accommodate vision-based tactile sensors while maintaining high manipulability. The approach generates URDF variants with different thumb parameterizations and evaluates them through collision checking, configuration filtering, and metric computation. For each URDF variant, we modify the thumb's distal joint by applying specified z-offset and pitch angle, loading the resulting URDF into Pinocchio for forward kinematics computation. Since our tactile sensor has a deformable gel surface, each fingertip's collision geometry is shrunk by approximately 15\% to account for small contact deformations.

\begin{figure}[t]
    \centering
    \includegraphics[width=0.49\textwidth]{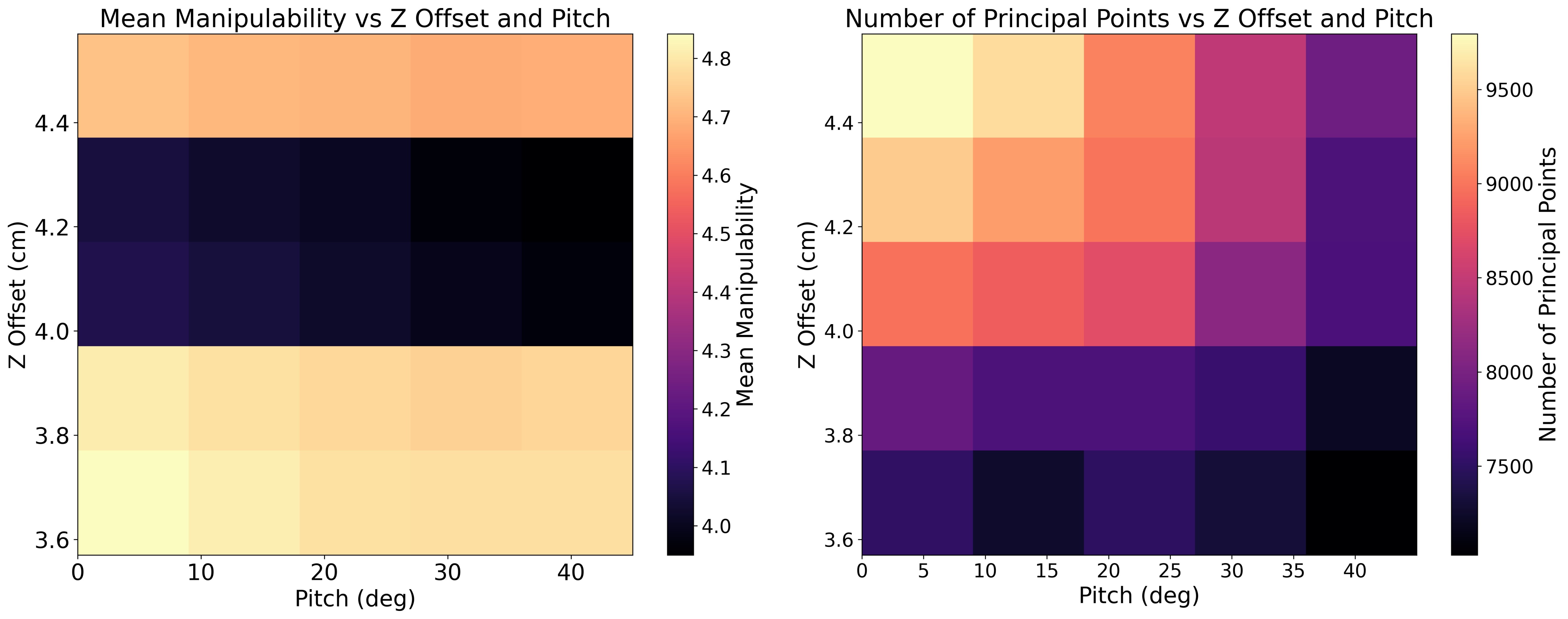}
    \caption{(Left) Mean manipulability $\bigl(\sqrt{\det(\mathbf{J}\mathbf{J}^\top)}\bigr)$ across different $z$-offsets (cm) and pitch angles (deg). (Right) The number of principal points (i.e., fingertip contact points inliers) for each $(z,\text{pitch})$ pair.}
    \vspace{-1.7em} 
    \label{fig:manipulability_vs_z_pitch}
\end{figure}

We create a grid of joint angles for both thumb and index fingers, then use forward kinematics to compute fingertip poses and Jacobians. Configurations with near-zero manipulability (below  $10^{-23}$) are discarded. Additional filters ensure contacts occur within 6 cm distance and require proper normal alignment between fingertips (opposing normals with negative dot product). For viable configurations, we query the collision detection system for exact contact points between sensor surfaces. These contact points represent physically feasible thumb-index contacts and form our core dataset for each URDF variant, as shown in Fig. \ref{fig:principal_axis_and_best_inlier_axis}.

To measure contact point alignment, we employ a RANSAC-like cylinder-fitting procedure that searches for a 3D axis maximizing the number of contact points within 1 cm radius. This identifies a ``best inlier axis'' along which fingers can most reliably make contact. We also compute a simpler principal axis using PCA on the contact point cloud.

Fig. \ref{fig:manipulability_vs_z_pitch} presents heatmaps summarizing mean manipulability and the number of well-aligned contact points for different parameter combinations. Although some configurations yield higher manipulability alone, we sought balance between dexterity and contact reliability. The variant $z=4.3\,\mathrm{cm}$, pitch $=0.0^\circ$ are selected as optimal, which exhibits numerous well-aligned contact points while maintaining strong overall manipulability.
The cylinder-fitting results in Fig. \ref{fig:principal_axis_and_best_inlier_axis} reveal how effectively contact points cluster along a single axis, demonstrating the method's ability to identify configurations suitable for reliable pinching operations.

\bibliographystyle{IEEEtran} % Specifies the bibliography style
\bibliography{ref}

% \newpage

 

% \end{IEEEbiography}



% \vfill

\end{document}